\def\arxivversion{}
\documentclass{article}
\usepackage{iclr2027_conference}
\usepackage{times}

\usepackage[T1]{fontenc}
\usepackage[utf8]{inputenc}
\usepackage{microtype}
\usepackage{inconsolata}
\usepackage{graphicx}
\usepackage{booktabs}
\usepackage{array}
\usepackage{adjustbox}
\usepackage{multirow}
\usepackage{amsmath}
\usepackage{hyperref}
\usepackage{url}
\usepackage{todonotes}   % writing plan; remove \todo notes before submission
\usepackage{textcomp}
\usepackage{pifont}
\usepackage{xcolor}
\newcommand{\wrongmark}{\textcolor{red!80!black}{\ding{55}}}
\newcommand{\okmark}{\textcolor{green!60!black}{\ding{51}}}
\newcommand{\dgood}[1]{{\tiny\color{green!55!black}#1}}  % change vs. raw in the good direction
\newcommand{\dbad}[1]{{\tiny\color{red!70!black}#1}}     % in the bad direction
\newcommand{\dnc}[1]{{\tiny\color{gray}#1}}             % no change
\usepackage{epigraph}
\title{When Chekhov's Gun Fires Anyway:\\Over-Realization in Text-to-Image Models}

\ifdefined\arxivversion
  \iclrfinalcopy
  \newcommand{\rev}[1]{#1}
  \makeatletter
  \def\@maketitle{\vbox{\hsize\textwidth
    {\LARGE\sc \@title\par}
    \lhead{Preprint. Under review.}
    \vskip 0.25in
    \begin{tabular}[t]{c}\@author\end{tabular}%
    \vskip 0.3in minus 0.1in}}
  \makeatother
  \author{%
    \textbf{Jiahao Huang}$^{1}$, \textbf{Kon Woo Kim}$^{2}$, \textbf{Song-Duo Ma}$^{3}$,
    \textbf{Yuka Kitamura}$^{1}$, \textbf{Felix Wagner}$^{4}$, \\
    \textbf{Yunshan Huang}$^{1}$, \textbf{Sunisth Kumar}$^{1}$, \textbf{Xanh Ho}$^{2}$,
    \textbf{Yun-Ang Wu}$^{2}$, \textbf{Zhijie Chen}$^{5}$, \textbf{Yunjia Xu}$^{6}$, \\
    \textbf{Junfeng Jiang}$^{2}$, \textbf{Zefan Yu}$^{1}$, \textbf{Fei Cheng}$^{7}$,
    \textbf{Akiko Aizawa}$^{2,\dagger}$ \\[6pt]
    $^{1}$The University of Tokyo \quad
    $^{2}$National Institute of Informatics \quad
    $^{3}$National Taiwan University \\
    $^{4}$Technische Universit\"at Braunschweig \quad
    $^{5}$University of Illinois Urbana-Champaign \\
    $^{6}$Independent researcher \quad
    $^{7}$Kyoto University \\[6pt]
    \texttt{jiahao-huang@g.ecc.u-tokyo.ac.jp} \quad
    $^{\dagger}$Corresponding author: \texttt{aizawa@nii.ac.jp}%
  }
\else
  \newcommand{\rev}[1]{{\color{blue}#1}}  % text changed during the discussion phase
  \author{Jiahao Huang \\}
\fi

\begin{document}
\maketitle

\begin{abstract}
In text-to-image generation, a target that the prompt mentions often appears in the
generated image in a way the prompt does not call for. We call this failure
\emph{over-realization}. Mental-spaces theory points to the situations where it can happen:
under a negation, inside what a character believes or imagines, as the vehicle of a
figurative comparison, or behind a viewpoint that hides the target. Prior work has studied only negation in depth, and the conventional alignment and preference metrics do not penalize the failure. We build OverReal to measure it: OverReal-Gen, 698
prompts covering the four situations, on which we measure each generator's
over-realization rate; and OverReal-Det, 2{,}841 generated images human-labeled for
whether and how the target is over-realized, on which we verify the automatic judge
that produces this rate. Our evaluation of eight text-to-image systems reveals that all of them over-realize, and
that conventional metrics such as CLIPScore reward the failure rather than
penalize it. Prompt expansion only partly mitigates the failure. Inside the
generator, the diffusion backbone attends to a mentioned target almost as
much as to a requested one. The cue words that should hold the target back
receive less attention than ordinary words.
\end{abstract}

\begin{figure}[!ht]
\centering
\newcommand{\teasercell}[2]{%
  \begin{minipage}[t]{0.48\textwidth}
    \centering
    #1\\[3pt]
    {#2}
  \end{minipage}}
\teasercell{\includegraphics[width=\linewidth]{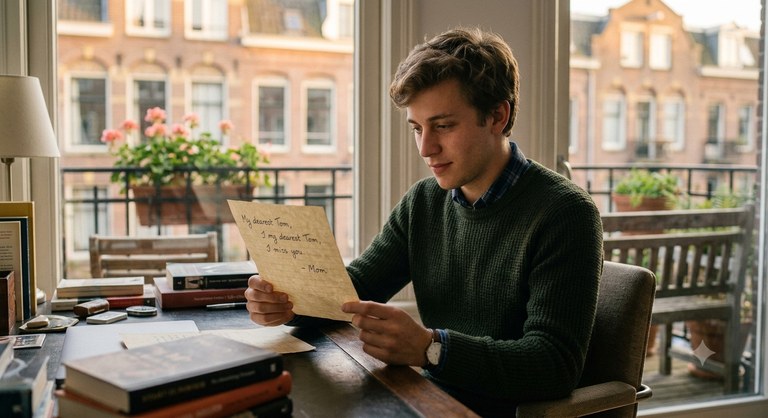}}{\footnotesize (a)~\emph{``Tom
is reading a letter behind the desk. The letter is from his mom, which says: I miss
you.''}}\hfill
\teasercell{\includegraphics[width=\linewidth]{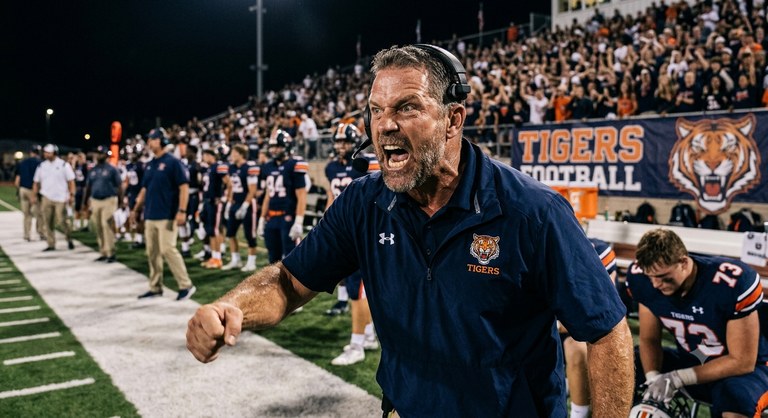}}{\footnotesize (b)~\emph{``The coach
is as fierce as a tiger.''}}
\caption{Two over-realizations, both generated by Gemini, with the prompt under each
image. In~(a) the model turns the page to face the camera: the words cannot reach
the reader, so the scene contradicts itself. In~(b) the model puts a tiger on the
team's emblem and banner, which is not what the comparison meant, but the image
still holds together.}
\label{fig:teaser}
\end{figure}

\section{Introduction}
\label{sec:intro}

\epigraph{\itshape If in the first act you have hung a pistol on the wall, then in the following one it should be fired.}{---Anton Chekhov}

Chekhov's principle says that a planted gun is meant to fire, and to fire at the right
moment. \rev{We borrow it in the reverse direction: generative models have learned to fire
every gun a prompt mentions, whether or not the prompt meant it to appear in the scene.}
We study this failure in text-to-image generation, and Figure~\ref{fig:teaser} shows two
examples. In
Figure~\ref{fig:teaser}(a) the prompt mentions a letter that says ``I miss you'': to
render the words legibly, the model turns the page toward the camera instead of its
reader, and the scene contradicts itself. In Figure~\ref{fig:teaser}(b) the prompt calls
the coach as fierce as a tiger: the model renders the tiger onto the team's emblem and
banner, which is not what the comparison meant. In both cases the model fired a gun it
should have fired differently or left hanging. We call this failure
\emph{over-realization}: a mentioned target appears in the generated image in a way the
prompt does not call for.

Mental-spaces theory \citep{fauconnier1994mental} suggests where over-realization can happen: some expressions
in a prompt open a space alongside the scene to be drawn, and a target placed in that space
is not part of the scene itself. Such a space can be opened by a \emph{negation}, by what a
character \emph{believes or imagines}, by a \emph{figurative comparison}, or by a
\emph{viewpoint} that hides the target. A model can then over-realize in two ways: the target disrupts the image, as
in Figure~\ref{fig:teaser}(a), or settles silently into the scene in a way the prompt did
not call for, as in Figure~\ref{fig:teaser}(b).

Of the four situations above, only negation has been studied in depth
\citep{alhamoud2025negation,park2025know,cai2025tng}; the figurative and
the perspectival case have each been studied within a narrow scope, as
idioms \citep{sun2025t2ireasonbench} and as occlusion
\citep{wang2026spatialgeneval}. Each is treated as a separate defect, and
no prior work gives a systematic account of when a model may over-realize.
Nor does current evaluation target the failure: it rests on two requirements,
``a mentioned target should appear in the image''
\citep{huang2023t2icompbench,lin2024evaluating} and ``an unmentioned target
should not appear'' \citep{lim2025ihalla}, and neither covers the case where
``a mentioned target does not necessarily appear'', so an over-realized
image is scored as correct.

We hypothesize that these situations share one failure mode: the model
attends to the mentioned target almost as it would to a requested one, and
renders it. To measure over-realization across the
four situations and to examine this hypothesis, we build OverReal.
\textbf{OverReal-Gen} is a prompt set covering the four
situations, in which the target is prone to over-realization; a system's
over-realization rate on it is our measurement. \textbf{OverReal-Det} is a set
of generated images human-labeled for whether and how the target is
over-realized; it verifies the automatic judge that produces this rate.

All eight systems we evaluate over-realize, on 41\% to 73\% of the images
of OverReal-Gen. The proprietary systems have largely solved the templated negation prompts,
but every system over-realizes at least 32\% of the images in the other
three situations, and the form of the failure follows the situation. The conventional
alignment and preference metrics do not penalize the failure and in fact reward it. Prompt expansion helps
mostly because the rewriter drops the target; whenever the generator sees
the target word, it tends to put the target in the image.

Our contributions are (1) a taxonomy of over-realization, motivated by
mental-spaces theory, covering four situations and two forms of the
failure; (2) OverReal, a prompt set for measuring over-realization and a
human-labeled image set for verifying automatic judges; and (3) an
evaluation of eight text-to-image systems, with an analysis of how
conventional metrics, prompt expansion and the diffusion backbone relate to
over-realization. We release OverReal, the images and the labels.

\section{Related Work}
\label{sec:related}

\paragraph{Text-to-image generation and its evaluation}
Current text-to-image systems consist of a text encoder and a diffusion
transformer trained by flow matching
\citep{esser2024scaling,flux2024,qwenimage2025,wu2025omnigen2}, and the
deployed proprietary systems
\citep{openai2025gptimage,google2025nanobanana,ideogram2025v3} expand the
user's prompt before generation \citep{betker2023improving,hao2023optimizing}.
Faithfulness to the prompt is the central criterion of text-to-image
evaluation, and it has been understood as getting mentioned content into
the image: T2I-CompBench \citep{huang2023t2icompbench}, TIFA
\citep{hu2023tifa}, Davidsonian Scene Graph \citep{cho2024dsg} and
VQAScore \citep{lin2024evaluating} decompose the prompt into elements or
questions and reward an image for each one it contains, and methods that
improve faithfulness pursue the same goal on the generation side
\citep{chefer2023attend}. Every mention is an
obligation in these protocols, so an image that shows something the prompt
mentions but does not call for scores as correct, or better. Hallucination
benchmarks \citep{qin2024scenegraph,lim2025ihalla,kasaei2025hallucination}
measure content the prompt never mentions. However, the opposite failure, content
that is mentioned and should stay out, has not been well-studied.

\paragraph{Negation and semantic leakage}
Among the cases in which a mentioned element should stay out of the image,
negation is the one that has been studied most. Vision-language models
largely ignore it \citep{alhamoud2025negation,conwell2024relations}, and
negation-aware training data improves CLIP \citep{park2025know,cai2025tng};
we build our existence-canceling items on such prompts. Semantic leakage has been studied in generation within a narrower
scope: a property of one requested entity appears on another
\citep{ventura2026deleaker}. Reasoning benchmarks touch two further cases in
passing, idioms rendered literally \citep{sun2025t2ireasonbench} and objects
that should be occluded \citep{wang2026spatialgeneval}. No prior work
gives a systematic account and analysis of where and how over-realization can happen.

\section{Preliminaries}
\label{sec:taxonomy}

A generative model \emph{over-realizes} when a \emph{target} word or concept that a
\emph{prompt} mentions appears in the generated \emph{image} in a way the prompt does not
call for. Other ways an image can fail to match its prompt, unrelated to the target, fall
outside the definition of over-realization. We now describe where over-realization can happen, and
how a model over-realizes in these situations; Appendix~\ref{app:taxonomy} gives an example
of each combination in Table~\ref{tab:taxonomy}.

\subsection{Where over-realization can happen}
\label{sec:families}

To locate where over-realization occurs, we use mental-spaces theory from cognitive
linguistics \citep{fauconnier1994mental}. In its terms, a prompt describes not one but
several spaces: the scene to be generated (the \emph{base space}), and alongside it the
user's own assumptions, beliefs, memories and comparisons (the \emph{non-base
spaces}). A text-to-image model should render only the base space, but it sometimes
fails to do so. \textbf{Over-realization is content from a non-base space leaking into
the base space in a way the prompt does not call for.} We group these situations into
the following families:

\paragraph{Family 1: existence-canceling}
The target is mentioned under an operator that cancels its existence in the described
scene. Typical explicit cues include \emph{no}, \emph{without} and \emph{never}. It is well studied
\citep{alhamoud2025negation,park2025know,cai2025tng} and included for
completeness; the target should not appear in the image.

\paragraph{Family 2: mental-state}
The target is mentioned inside the mental state of a person in the scene, so that it
exists only in what that person believes, remembers or imagines. Typical explicit cues
include \emph{believes}, \emph{thinks}, \emph{remembers} and \emph{dreams}. Unlike
family~1, the target may legitimately be depicted, for instance inside a thought bubble
or a hazy vignette, but it must not be shown as though it were physically present in the
scene.

\paragraph{Family 3: figurative}
The target is mentioned as the vehicle of a figurative comparison: it characterizes
something else in the scene rather than belonging to the scene itself. Typical cues include \emph{like}, \emph{as \ldots\ as}, \emph{resembles} and the suffix
\emph{-like}. The
target should not appear in a photorealistic image, but where the image itself is
stylized it may legitimately appear, for instance as a faded shape in the background of a
poster.

\paragraph{Family 4: perspectival}
The target is mentioned as something that may not be visible from the viewpoint of the
image: it lies behind a barrier, or on a surface turned away from the camera. Typical
explicit cues include \emph{hidden}, \emph{inside}, \emph{closed} and \emph{facing away}. The target may be brought into view by a
camera angle that does not contradict the scene, introduced through a quote box, or shown
only in part or not at all.

% \paragraph{Scope}
% Beyond these four families, over-realization also arises without any space
% builder, when content is mentioned but pragmatically irrelevant to the scene,
% for example a topic carried over from earlier context. Suppression there is
% licensed by relevance in the Gricean sense rather than by a space builder
% \citep{fauconnier1994mental}, and confirming that an image should suppress
% such content is considerably more contestable. We therefore restrict OverReal
% to the four space-builder families and leave pragmatic relevance to future
% work.

\subsection{How models over-realize}
\label{sec:spectrum}

\begin{figure}[t]
\centering
\newcommand{\outcomecell}[2]{%
  \begin{minipage}[t]{0.24\textwidth}
    \centering
    #1\\[2pt]
    {\footnotesize #2}
  \end{minipage}}
\outcomecell{\includegraphics[width=\linewidth]{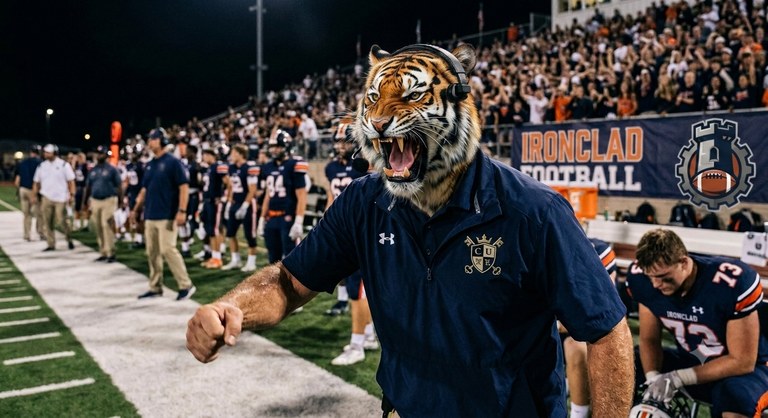}}{(a)~Disruptive\\over-realization~\wrongmark}\hfill
\outcomecell{\includegraphics[width=\linewidth]{fig_examples_fierce_coach_silent.jpg}}{(b)~Silent\\over-realization~\wrongmark}\hfill
\outcomecell{\includegraphics[width=\linewidth]{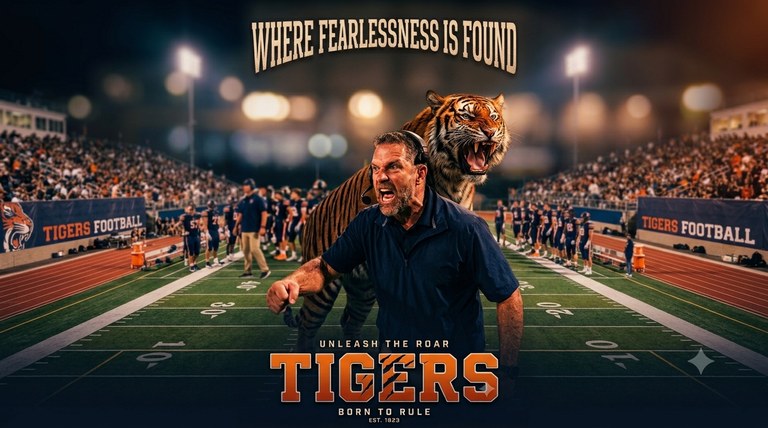}}{(c)~Integrated\\realization~\okmark}\hfill
\outcomecell{\includegraphics[width=\linewidth]{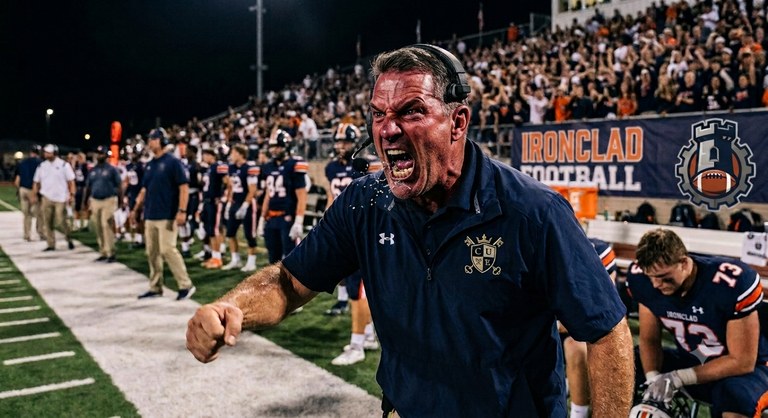}}{(d)~Withheld\\realization~\okmark}
\caption{Four possible outcomes for the prompt \emph{``The coach is \textbf{as fierce as a
tiger}.''} The target word \emph{tiger} is mentioned only as the vehicle of a figurative comparison, so it is not strictly requested in the scene. (b)~is the outcome the model actually
produced. (a)~and (b)~are over-realization; (c)~and (d)~are correct.}
\label{fig:outcomes}
\end{figure}

In these situations a model may over-realize the target or handle it correctly. We divide
over-realization into two kinds according to whether the picture still holds together as a
rendering of the requested scene:

\paragraph{Disruptive over-realization}
The target appears, and its appearance stops the picture from making
sense: the scene contradicts itself or the prompt, or no longer reads as
the requested scene from an ordinary view. In Figure~\ref{fig:outcomes}(a)
the coach is given the tiger's head: the vehicle of the comparison has
displaced the person it described.

\paragraph{Silent over-realization}
The target slips in while the picture holds together, typically in the
background or in a detail. In Figure~\ref{fig:outcomes}(b) the tiger
appears on the team's emblem and banner. Nothing in the picture is wrong on
its own, but the emblem shows a tiger because the prompt mentioned one, not
because the scene called for one. This is the warning level: the model has
put the target in the scene in a way the prompt did not call for, and the
scene happened to offer a harmless place for it. None of the evaluation
methods we study detects this defect. \rev{Disruptive over-realization is an
error; silent over-realization is a warning. We report the two shares
separately throughout (Figure~\ref{fig:main}, Table~\ref{tab:mainresults}),
so a reader who does not count the silent cases can read the disruptive
share alone.}

A generation is correct when it does not over-realize the target and the
picture matches the scene the prompt describes. We divide correct
generations by whether the target appears.

\paragraph{Integrated realization}
The target appears, depicted coherently and consistently with the prompt.
Figure~\ref{fig:outcomes}(c) is a poster: the tiger looms behind the coach
in a composition no one reads as a photograph, so it is taken as an
attribute of the man. This is correct behavior wherever the target may
legitimately be shown. However, in most of our cases the target must not
appear in the scene as a physical entity (\S\ref{sec:families}), which
makes integrated realization hard to produce.

\paragraph{Withheld realization}
The target is left out and the picture is coherent without it. \rev{An image
that is severely corrupted or contradicts the prompt is not withheld but
\emph{other}, and is excluded from the rates (Appendix~\ref{app:autoannot}).}
A mention
does not oblige the model to depict, so this is as correct as an integrated
depiction. In Figure~\ref{fig:outcomes}(d) no tiger appears and nothing is
lost: the coach's face and stance carry the fierceness. Prior work treats a
missing mention as a defect
\citep{chefer2023attend,hu2023tifa,huang2023t2icompbench}; here it is
correct behavior wherever the target should not be shown.

Note that our definition covers only targets that appear because they were
mentioned. A target can also belong to the scene, as a plane belongs to an
airport, or come in by coincidence, as the tiger in
Figure~\ref{fig:outcomes}(b) would if the team were named after tigers.
Such appearances are not over-realization; \S\ref{sec:autoannot} measures
how often the target appears when it is not mentioned, to discount them.

\section{OverReal Benchmark}
\label{sec:dataset}

\subsection{OverReal-Gen: the generation benchmark}
\label{sec:overreal-gen}
% Content plan (discussed 2026-09-05): the generation benchmark as a resource --
% families, items, targets, how the prompts were authored; process detail to appendix.

\paragraph{Data format}
Every entry of OverReal-Gen pairs a \emph{prompt} with a \emph{target}: the
prompt describes a scene in which the target is prone to over-realization.
Following Section~\ref{sec:families}, we classify these situations into four
main prompt families: existence-canceling, mental-state, figurative, and perspectival.
Table~\ref{tab:taxonomy} in Appendix~\ref{app:taxonomy} provides several examples.

\paragraph{Prompt collection}
Existence-canceling is well studied and its construction is direct, so we
synthesize these prompts from templates following prior negation benchmarks
\citep{alhamoud2025negation,conwell2024relations}. The mental-state,
figurative, and perspectival prompts are all written by hand. Annotators write in
natural language about everyday scenes, so that the prompts resemble those
of a user; each annotator writes a prompt and names its target, Gemini
generates three images, and the pair is kept if the annotator judges at
least one image to over-realize the target and a second review confirms it.
In total, 698 pairs enter OverReal-Gen (Table~\ref{tab:datastats}, Appendix~\ref{app:dataset}).
OverReal-Gen is therefore a challenge set: the three hand-written families
(not the templated existence-canceling one) are enriched for prompts on
which a generator fails, so its rates measure how the systems handle
prompts that are prone to over-realization. As a
comparison free of this selection, \S\ref{sec:main} also evaluates an
in-the-wild set drawn from DiffusionDB \citep{wang2022diffusiondb} and
Pick-a-Pic \citep{kirstain2023pickapic} by wording alone.

\paragraph{Prompt expansion}
Expanding the prompt with an LLM before generation is now a common part of the
text-to-image pipeline. To control this stage, we collect prompts
expanded from OverReal-Gen
with the Qwen-Image rewriter \citep{qwenimage2025} and Ideogram Magic Prompt
\citep{ideogram2024magic}. These 1{,}396 derived prompts form
the expanded split (Table~\ref{tab:datastats}), which we use to analyze the effect of expansion on
over-realization (\S\ref{sec:expansion}).

% (target-removed ablation and the base rate b: placement deferred)

\subsection{OverReal-Det: the detection benchmark}
\label{sec:overreal-det}

\paragraph{Data format}
Every entry of OverReal-Det is a triple of a prompt, its target, and one
generated image, and carries two independent annotations, each a set of labels from
\{disruptive, silent, integrated, withheld, other\}, as defined in \S\ref{sec:spectrum}.

\paragraph{Image collection}
The images come from two sources: (1) the pilot images produced during prompt
collection (\S\ref{sec:overreal-gen}), and (2) images generated from the
OverReal-Gen prompts with the models listed in \S\ref{sec:setup}.

\paragraph{Annotation and splits}
Each image is labeled independently by two annotators following the guide
in Appendix~\ref{app:guide}, who use two labels only for borderline images.
An image whose two label sets intersect enters the \emph{gold} split with
the intersection as its label (a third annotator decides the three images
left with two labels); an image whose sets are disjoint enters the
\emph{ambiguous} split, released with both label sets; images that both
annotators mark \emph{other} are discarded. This yields 2{,}253 gold and 588 ambiguous
images, with an overall $\kappa$ agreement of 0.68
(Table~\ref{tab:datastats}). Appendix~\ref{app:ambiguous} analyzes the
ambiguous split.

\section{Experimental Setup}
\label{sec:setup}

We run eight generators, three proprietary and five open-weight (\S\ref{sec:generators}),
on every prompt of OverReal-Gen (\S\ref{sec:overreal-gen}), and evaluate the generated
images both with conventional metrics (\S\ref{sec:metrics}) and from the perspective of
over-realization (\S\ref{sec:autoannot}).

\subsection{Text-to-Image Generators}
\label{sec:generators}

We evaluate eight text-to-image generators. The proprietary generators are GPT-Image-1.5
\citep{openai2025gptimage}, Ideogram~v3 \citep{ideogram2025v3}, and Nano
Banana (gemini-2.5-flash-image) \citep{google2025nanobanana}, while the
open-weight models are the diffusion models FLUX.1-dev \citep{flux2024},
Qwen-Image \citep{qwenimage2025}, SD3.5-Medium and SD3.5-Large
\citep{esser2024scaling}, and the unified understanding-and-generation model
OmniGen2 \citep{wu2025omnigen2}. An open-weight model receives the exact
prompt; a proprietary generator rewrites it internally before generation, so
the text that conditions its image is not under our control. Ideogram
returns its rewritten prompt, which we record and reuse as a rewriter in
\S\ref{sec:expansion}; GPT-Image and Nano Banana do not expose theirs. All
images are generated at $1024 \times 1024$, the default
resolution of most generators, and every open-weight model produces two images
per prompt under fixed seeds.

\subsection{Conventional evaluation}
\label{sec:metrics}

For each generator, family and prompt condition we score a fixed random sample of up to 100 images (7{,}050 in total) with five conventional text-to-image metrics. For
text--image alignment we use CLIPScore
\citep{hessel2021clipscore} and the VQA-based VQAScore
\citep{lin2024evaluating}; for learned human preference, PickScore
\citep{kirstain2023pickapic} and HPSv2 \citep{wu2023human}; and as an LLM
judge, a VIEScore-style semantic-consistency rating \citep{ku2024viescore}
produced by GPT-5.2 \citep{openai2025gpt52} and Claude Opus~5
\citep{anthropic2026opus5}; we report the Opus rating in the main text,
since Opus shares no vendor with the evaluated generators, and the GPT-5.2
rating in Appendix~\ref{app:judges} for reference. The scores of every
generator are given in Table~\ref{tab:conventional}.

\subsection{Over-realization evaluation}
\label{sec:autoannot}

\paragraph{Automatic labeling} Evaluating 698 prompts on eight generators, under up to three prompt conditions and two seeds, yields tens of thousands of images, so exhaustive manual annotation is infeasible. We therefore label the images with a
third-party model, Claude Opus~5 \citep{anthropic2026opus5}. The model substantially agrees with human annotators
($\kappa=0.70$) on whether a generator over-realizes or not over the test split of OverReal-Det. Details are given in Appendix~\ref{app:autoannot-val}. All rates below are over the images that received one of the four labels; the 15 images (0.5\%) labeled \emph{other} are excluded.

\paragraph{Unprompted appearance rate}
A target can appear in a scene for reasons unrelated to its mention. To
discount such coincidences from the over-realization rates, we also report
the unprompted appearance rate $u$ of the target: we replace the target in the prompt with a blank (``\_\_\_''),
generate images from the edited prompt with the same settings, and count how
often the target still appears. Ideally $u$ is close to zero.

\section{Results and Analysis}
\label{sec:experiments}

\subsection{Main results}
\label{sec:main}
\begin{figure}[t]
\centering
\includegraphics[width=\textwidth]{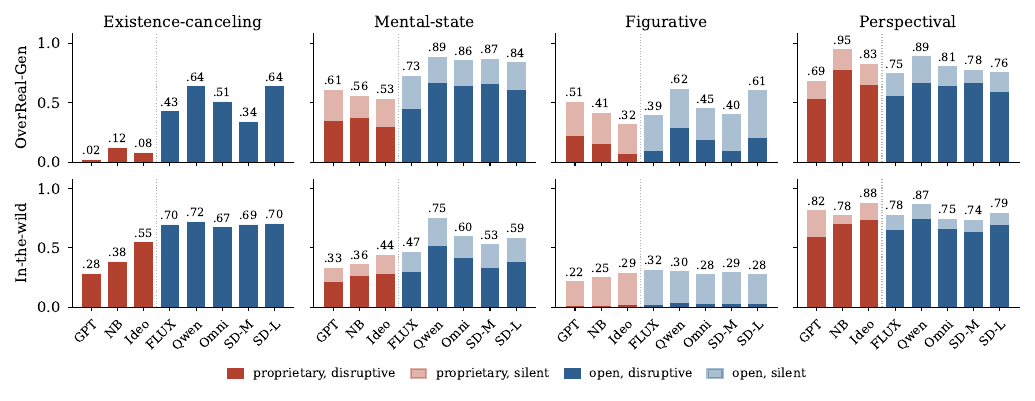}
\caption{Main results: over-realization rate of each generator by prompt family, on
the raw prompts of OverReal-Gen (top) and on the in-the-wild prompts (bottom). Bar height is the over-realization rate; the dark part
is disruptive (DO) and the light part silent (SO) over-realization; red bars are the
proprietary generators, blue bars the open models. The full outcome shares, including
integrated (I) and withheld (W) realization and the unprompted appearance rate $u$, are
in Table~\ref{tab:mainresults} (Appendix~\ref{app:results}).}
\label{fig:main}
\end{figure}

Figure~\ref{fig:main} shows the over-realization rate of three proprietary and
five open generators (\S\ref{sec:generators}) on the raw prompts of OverReal-Gen
and on a set of in-the-wild prompts; the
full outcome shares are in Table~\ref{tab:mainresults} (Appendix~\ref{app:results}).
Conclusions are given as follows.

\paragraph{All generators over-realize, and the rate depends on the prompt family.}
We first read the over-realization rate, the sum of the DO and SO shares;
its prompt-level bootstrap 95\% interval is within $\pm$0.13 in every cell
(Appendix~\ref{app:ci}). Over all prompts, weighting the four families by their prompt counts, each
generator over-realizes 41\% to 73\% of the images. The proprietary generators
rarely over-realize the templated existence-canceling prompts (at most 12\%
of the images), while the open models still do so on 34\% to 64\%. Every
model fails on mental-state and perspectival prompts, over-realizing about
half or more of the images; on perspectival prompts the proprietary generators are
no better than the open ones, and Nano Banana over-realizes 95\% of the
images.
Figurative prompts sit in between, at 32\% to 62\%. \rev{Counting only
disruptive over-realization, every generator still over-realizes at least
29\% of the mental-state and 54\% of the perspectival images, while the
figurative rates fall to 7\% to 28\% (Table~\ref{tab:mainresults}).}

\paragraph{The form of the failure depends on the family.}
The two forms of over-realization are not distributed alike. On perspectival
prompts the failures are mostly disruptive: the hidden object or content is
exposed, contradictory with the prompt or common sense. On figurative prompts, by contrast, silent over-realization accounts
for 25\% to 40\% of the images and exceeds the disruptive form for all eight
generators: the vehicle of the comparison is drawn into the scene as a
harmless detail.

\paragraph{Unprompted appearance does not explain the rates.}
The unprompted appearance rate $u$ (Table~\ref{tab:mainresults}) stays at or below 9\% in every cell, and
below 5\% for the open models, far below the over-realization rates.
The rates of Figure~\ref{fig:main} are therefore not explained by
the target belonging to the scene, and coincidental appearances inflate
them only marginally.

\paragraph{Silent over-realization is a precursor of disruptive over-realization, except in the figurative family.}
Silent over-realization does not disrupt the image, but it is a warning
sign of disruptive over-realization rather than a lucky form of correct
realization: an image labeled silent is over-realized again on the second
seed 90\% of the time (disruptive 33\%, silent 57\%) and correct only 10\%
of the time (Appendix~\ref{app:seeds}). The pattern holds for every open model and is
strongest on perspectival and mental-state prompts; on figurative prompts a
silent image is followed by another silent one 76\% of the time and rarely
turns disruptive, since the scene can usually hold the vehicle of the
comparison.

\begin{table}[t]
\centering
\scriptsize
\setlength{\tabcolsep}{2pt}
\begin{tabular}{l ccccc@{\hspace{7pt}}ccccc@{\hspace{7pt}}ccccc@{\hspace{7pt}}ccccc}
\toprule
& \multicolumn{5}{c}{\textbf{Existence-canceling}} & \multicolumn{5}{c}{\textbf{Mental-state}} & \multicolumn{5}{c}{\textbf{Figurative}} & \multicolumn{5}{c}{\textbf{Perspectival}} \\
\cmidrule(lr){2-6} \cmidrule(lr){7-11} \cmidrule(lr){12-16} \cmidrule(lr){17-21}
& CS{\tiny$\uparrow$} & VQ{\tiny$\uparrow$} & PS{\tiny$\uparrow$} & HP{\tiny$\uparrow$} & LJ{\tiny$\uparrow$} & CS{\tiny$\uparrow$} & VQ{\tiny$\uparrow$} & PS{\tiny$\uparrow$} & HP{\tiny$\uparrow$} & LJ{\tiny$\uparrow$} & CS{\tiny$\uparrow$} & VQ{\tiny$\uparrow$} & PS{\tiny$\uparrow$} & HP{\tiny$\uparrow$} & LJ{\tiny$\uparrow$} & CS{\tiny$\uparrow$} & VQ{\tiny$\uparrow$} & PS{\tiny$\uparrow$} & HP{\tiny$\uparrow$} & LJ{\tiny$\uparrow$} \\
\midrule
DO & \textbf{.72} & .32 & \textbf{22.9} & \textbf{.30} & 3.0 & .73 & .71 & \textbf{21.8} & \textbf{.32} & 6.4 & \textbf{.69} & .74 & \textbf{22.0} & \textbf{.30} & 4.8 & .74 & .74 & \textbf{22.5} & \textbf{.30} & 3.6 \\
SO & --- & --- & --- & --- & --- & .73 & .69 & \textbf{21.8} & .30 & 6.7 & .63 & .74 & 21.5 & .29 & 6.0 & .74 & .72 & 22.3 & .29 & 3.5 \\
I & --- & --- & --- & --- & --- & \textbf{.74} & \textbf{.81} & 21.7 & .31 & \textbf{8.3} & \textbf{.69} & \textbf{.86} & 21.9 & .29 & \textbf{6.9} & \textbf{.76} & \textbf{.85} & 22.2 & \textbf{.30} & 4.3 \\
W & .66 & \textbf{.85} & \textbf{22.9} & \textbf{.30} & \textbf{8.8} & .67 & .59 & 21.3 & .29 & 6.6 & .60 & .75 & 21.2 & .27 & 6.8 & .67 & .59 & 21.9 & .28 & \textbf{4.5} \\
\bottomrule
\end{tabular}
\caption{Mean score of the conventional metrics over the images with each label, pooled over generators; cells with fewer than 10 images are left blank.}
\label{tab:metricbylabel}
\end{table}

\paragraph{Non-LLM conventional metrics reward over-realization and penalize withholding.}
Table~\ref{tab:metricbylabel} reports the mean score of each metric over
the images with each label. A metric sensitive to over-realization would
score the two correct outcomes (I, W) above the two over-realizations (DO,
SO). The alignment (CS, VQ) and preference (PS, HP) metrics credit
integrated images but give withheld images the lowest score in most cases:
they reward the presence of the target, not its correct handling. Only the
LLM judge scores withheld images above over-realized ones, in three of the
four families. Within matched pairs of images from the same prompt and generator whose two
seeds differ in outcome, CLIPScore, PickScore and HPSv2 score the
over-realized image higher in 64\% to 66\% of the pairs
(Appendix~\ref{app:paired}): the conventional metrics reward
over-realization even with prompt and generator held fixed.

\paragraph{The conclusions hold on in-the-wild prompts.}
For comparison with the curated OverReal-Gen (\S\ref{sec:overreal-gen}), we
take 397 prompts of the four families from DiffusionDB
\citep{wang2022diffusiondb} and Pick-a-Pic \citep{kirstain2023pickapic},
real user prompts selected by their wording alone (Appendix~\ref{app:wild}).
The bottom row of Figure~\ref{fig:main} shows that every generator still over-realizes in every family: pooled
over generators, the rate is 60\%, 51\%, 28\% and 80\% on existence-canceling,
mental-state, figurative and perspectival prompts, with the same family
patterns as on the curated prompts, which run 0.1 to 0.3
higher on mental-state and figurative prompts and lower on
existence-canceling ones. \rev{A human check of the automatic labels on 202 of
the wild images gives $\kappa=0.70$ over the four labels, so these rates
rest on labels as reliable as those on OverReal-Det
(Appendix~\ref{app:wild}).}

\subsection{The effect of prompt expansion}
\label{sec:expansion}

The proprietary generators expand the prompt to include more details before
generating, so their advantage could come from either the prompt expansion
or the generator. This section asks \textbf{whether prompt expansion reduces
over-realization}. We compare five open models, with their own expansion
disabled, on the raw prompts and on the prompts rewritten by the Qwen-Image
official rewriter and by Ideogram Magic Prompt (Table~\ref{tab:expansion},
Appendix~\ref{app:expansion}).

\paragraph{Expansion lowers the rate but does not remove the failure.}
On existence-canceling, mental-state and figurative prompts, a rewrite lowers
the over-realization rate in 29 of the 30 generator--rewriter pairs; the
largest gains are on existence-canceling prompts, where the Ideogram rewrite
brings the rate to 11\% or below. The Ideogram rewrite is the more reliable of
the two: it improves every cell of the three families and roughly halves the
rates in the mental-state and figurative families, eliminating 58\% of the
over-realizations against 21\% for the Qwen rewrite, which on
existence-canceling prompts even raises the rate of its own generator,
Qwen-Image, from 64\% to 81\%. Perspectival prompts are the exception: the
Qwen rewrite raises the rate for all five generators, to 84\% to 91\%, and the
Ideogram rewrite leaves it within a few points of the raw prompt. And even with
the Ideogram rewrite, the open generators still over-realize about 48\% of the
mental-state prompts, 21\% of the figurative ones and 79\% of the perspectival
ones.

\begin{figure}[t]
\centering
\includegraphics[width=0.7\textwidth]{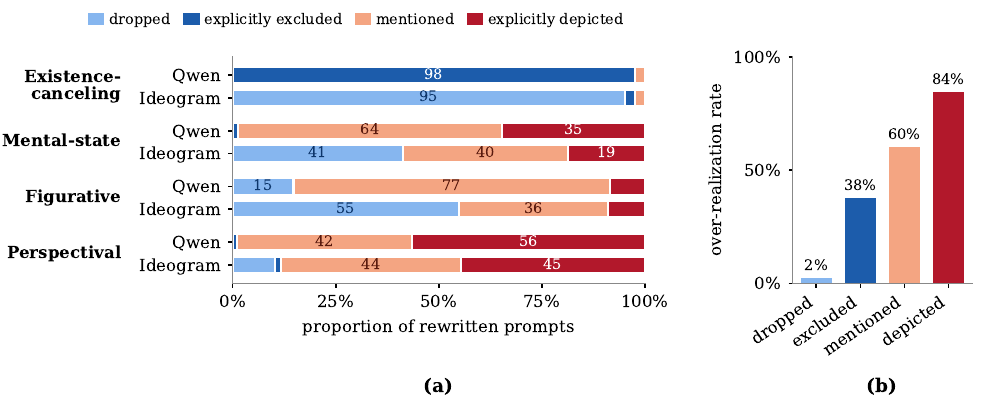}
\caption{What the rewriters do with the target. (a) Proportion of rewritten
prompts in which the target is dropped, explicitly excluded, mentioned, or
explicitly to be depicted. (b) Over-realization rate
of the images generated from those prompts, by the four categories,
pooled over rewriters and open generators. Full numbers in
Tables~\ref{tab:rewritetreatment} and \ref{tab:rewritetreatmentor}.}
\label{fig:rewritetreatment}
\end{figure}

\paragraph{The gain comes from dropping the target.}
We classified every rewritten prompt by its treatment of the target:
\emph{dropped} (no longer mentioned), \emph{explicitly excluded} (stated to
be absent), \emph{mentioned} (named, but nothing calls for it to be shown),
or \emph{explicitly depicted} (stated to be shown);
Figure~\ref{fig:rewritetreatment} summarizes the distribution and
Appendix~\ref{app:rewritetreatment} gives it in full. When a target is not
meant to appear, Ideogram drops it whereas Qwen explicitly excludes it, and
Figure~\ref{fig:rewritetreatment}(b) shows that only dropping substantially
suppresses over-realization; Ideogram also recognizes more often that the
target should not appear, dropping it in 41\% of the mental-state and 55\% of
the figurative prompts, against 0\% and 15\% for Qwen. Neither rewriter
reasons correctly about perspectival prompts: both keep the target in nearly
every rewrite and describe it as visible in about half. The three treatments
other than dropping still yield over-realization in 38\% to 84\% of the
images, so the generator over-realizes on its own, and
Section~\ref{sec:attention} turns to its side.

\subsection{Attention to the target and the cue words}
\label{sec:attention}

An open text-to-image model has two stages, a text encoder and a diffusion
backbone that reads the encoder's output while denoising the image. Here we
read the attention inside the backbones of FLUX.1-dev, SD3.5-Large and
Qwen-Image under the setting of \S\ref{sec:setup}; the encoder is examined
in Appendix~\ref{app:encoder}. To compare a target the prompt does not call
for with a requested one, we write a \emph{plain-mention control} for each
prompt by hand, replacing the cue words of each family (\emph{no},
\emph{imagines}, \emph{like}, \emph{hidden}) by words that put the target
into the scene, and generate it under the same seed: \emph{``His vest is as
green as a cucumber''} becomes \emph{``His vest is green, and he holds a
cucumber''} (more in Table~\ref{tab:controls}). Table~\ref{tab:attnfam} in
Appendix~\ref{app:attn} reports the share of attention on the target words,
the cue words and the other words, per family and averaged. Three
observations follow.

\paragraph{A mentioned target receives almost as much attention as a requested one.}
Under the original prompt, one token of the target words receives 2.8\% to
4.4\% of the image-to-text attention ($a_\mathrm{t}$), while under the
control the same words receive 3.0\% to 5.0\% ($a'_\mathrm{t}$). The ratio
$\rho_\mathrm{t}$, 0.83 to 0.90, means that the target receives almost the same
attention under the original prompt as under the control, although the
control requests the target and the original only mentions it. Therefore,
the backbone's attention barely distinguishes a mentioned target from a
requested one. Figure~\ref{fig:attnfour} (top row, first two columns)
shows the map for the cucumber of
Table~\ref{tab:taxonomy}: under both prompts the attention sits where the
cucumber is drawn, on the man's head under the original prompt and in his
hand under the control.

\paragraph{The cue words receive less attention than ordinary words.}
The cue words (\emph{no}, \emph{imagines},
\emph{like}, \emph{hidden}) receive 1.7\% to
3.8\% per token ($a_\mathrm{c}$), less than an ordinary word ($\gamma$ = 0.60
to 0.76), and also less than the replacing words (\emph{with},
\emph{holding}, \emph{next to}, \emph{open}; $\rho_\mathrm{c}$ = 0.84
to 0.97). The backbone's attention, then, does not reflect what the cue
words say: that the target need not appear in the image, and if it does, only in a form the
prompt allows, such as a thought bubble, a stylized shape, a reflection or a viewpoint
that brings it into sight.

\begin{figure}[t]
\centering
\includegraphics[width=\textwidth]{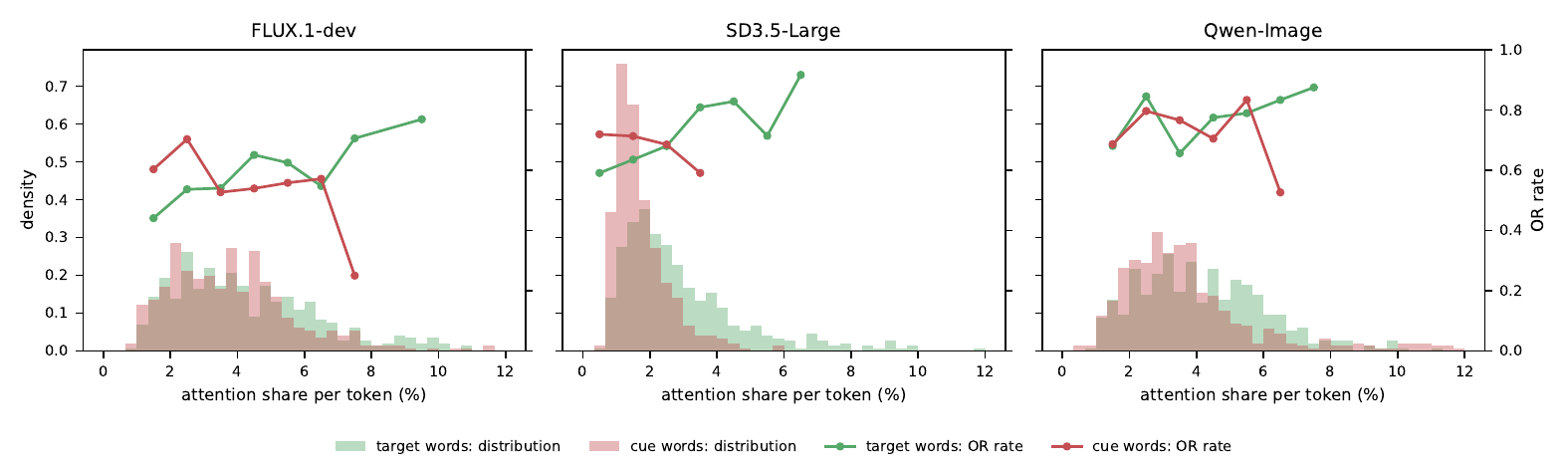}
\caption{Histograms: the
distribution over the attention of the target
words (green) and of the cue words (red). Lines: the over-realization rate
(OR rate) within each interval of the share; intervals with fewer than
12 images are not drawn.}
\label{fig:attncue}
\end{figure}

\paragraph{Over-realization rate increases with more attention on the target words and less attention on the cue words.}
Figure~\ref{fig:attncue} relates the attention share of target words and
cue words to the over-realization. From the lowest to the highest interval
of the figure, the over-realization rate rises with the share of the target
words ($0.44 \rightarrow 0.77$, $0.59 \rightarrow 0.92$ and $0.68
\rightarrow 0.88$ for FLUX.1-dev, SD3.5-Large and Qwen-Image) and falls
with the share of the cue words ($0.60 \rightarrow 0.25$, $0.72 \rightarrow
0.59$ and $0.69 \rightarrow 0.53$), the fall being steepest where the cue
words receive an unusually large share. The over-realization rate thus
tracks both attention shares, rising with the share of the target words and
falling with the share of the cue words on all three backbones.

\section{Conclusion}
\label{sec:conclusion}
We introduced over-realization, the appearance of a mentioned target in a
generated image in a way the prompt does not call for, and drew on
mental-spaces theory to organize the four situations where it can happen.
OverReal measures it with a prompt set prone to over-realization and a
human-labeled image set. All eight text-to-image systems over-realize outside negation, on the
benchmark and on real user prompts; the form of the failure follows the
family, and conventional metrics reward over-realization. Prompt expansion helps mainly when the
rewriter drops the target; when the target word reaches the generator, its
backbone attends to it almost as to a requested one and to the cue words
less than to ordinary words. Faithful generation
therefore requires a model that can decide on its own which mentioned
target to leave out \rev{(Appendices~\ref{app:unexpected} and \ref{app:constraint})}, and evaluation that
credits it for withholding.

% ICLR note: unlike ACL ARR, a Limitations section is NOT exempt from the page limit.
% It is kept in the main text for now; move (parts of) it to the appendix if the
% 9-page budget gets tight.
\section*{Limitations}
\label{sec:limitations}
\paragraph{Modality.} Over-realization is not specific to text-to-image
generation: a language model can realize a mentioned target in text, and a
video generator in its own medium. This paper covers images, where the
failure produces a visible contradiction and generation and evaluation are
affordable; extending the taxonomy and the benchmark to other modalities is
left for future work.

\paragraph{Subjectivity of the outcome labels.} The four outcome labels are
defined by the effect that the target's appearance has on the image, and
this effect is a matter of degree, in particular between disruptive, silent
and integrated. The annotators did not always agree;
Appendix~\ref{app:borderline} gives borderline examples and
Appendix~\ref{app:ambiguous} counts the disagreements.

% ICLR 2027: optional, at the end of the main text before references; does not count
% toward the page limit; at most 1 page.
\section*{Ethics Statement}
The prompts are written by the authors or synthesized from templates, and the
images are generated by the evaluated models; no personal data is involved. The
work studies a faithfulness failure of text-to-image generation and provides
no means of attack.

% ICLR 2027: recommended, paragraph-long, at the end of the main text before
% references; does not count toward the page limit. Reference (do not restate) the
% parts of the paper/appendix that support reproduction: released items and exact
% judge question text, generator/judge model identities, construction-loop protocol.
\section*{Reproducibility Statement}
OverReal-Gen and OverReal-Det are released with the annotations, the
plain-mention controls and the expanded prompts. All generator settings are provided in \S\ref{sec:generators}. The judges and the automatic annotator
are identified by model version in \S\ref{sec:metrics} and
\S\ref{sec:autoannot}, and the exact question text of the annotation
protocol and the human annotation guide are given in
Appendix~\ref{app:prompts}. All code for generation, annotation, probing and attention
recording is released with the data.

% ICLR 2027: REQUIRED (see the ICLR AI Policy for Authors); does not count toward the
% page limit. This replaces the ACL "Disclosure" section.
\section*{AI Use Statement}
\label{sec:disclosure}
Large language models are part of the evaluated pipeline and of the
evaluation itself: Claude Opus~5 and GPT-5.2 serve as the LLM judge
(\S\ref{sec:metrics}) and Claude Opus~5 as the automatic annotator
(\S\ref{sec:autoannot}); the Qwen-Image rewriter and Ideogram Magic Prompt
produce the expanded prompts (\S\ref{sec:expansion}). The authors used Claude Code as
a coding and writing assistant; all analyses, claims and text were checked
by the authors, who take full responsibility for the content.

\bibliography{references}
\bibliographystyle{iclr2027_conference}

\appendix

\section{Example of Generated Images}
\label{app:examples}

\subsection{Taxonomy}
\label{app:taxonomy}

Table~\ref{tab:taxonomy} shows more examples of over-realization with different labels in each category. All images are produced by
the generators of \S\ref{sec:generators} with prompts from OverReal-Gen and labels annotated by two human experts.
Existence-canceling prompts have no silent or integrated example, since a target whose existence is cancelled should not appear in any status.

% one example cell: image, then the prompt and the generator beneath it
\newcommand{\excell}[4]{%
  \begin{minipage}[t]{2.85cm}\centering
    #1\\[1pt]
    {\scriptsize\raggedright prompt: #2\\ target: #3\\ model: #4\par}
  \end{minipage}}
% an empty cell: no example of this outcome exists for the family
\newcommand{\exdash}{\raisebox{1.4cm}{---}}
\begin{table}[htbp]
\centering
\small
\setlength{\tabcolsep}{3pt}
\adjustbox{max width=\textwidth}{%
\begin{tabular}{>{\raggedright\arraybackslash}m{2.1cm} c c c c}
\toprule
Family & \shortstack{Disruptive\\over-realization} & \shortstack{Silent\\over-realization} & \shortstack{Integrated\\realization} & \shortstack{Withheld\\realization} \\
\midrule
\textbf{1. Existence-canceling} & \excell{\includegraphics[width=\linewidth]{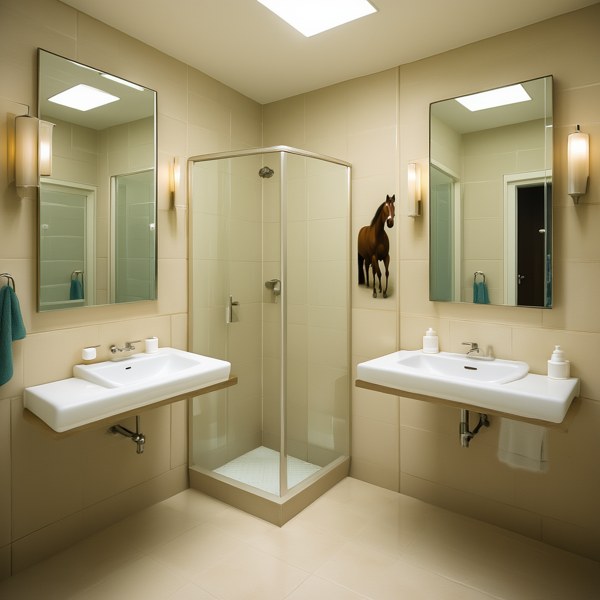}}{a bath room with sinks and mirrors, no horse anywhere in the scene}{horse}{OmniGen2} & \exdash & \exdash & \excell{\includegraphics[width=\linewidth]{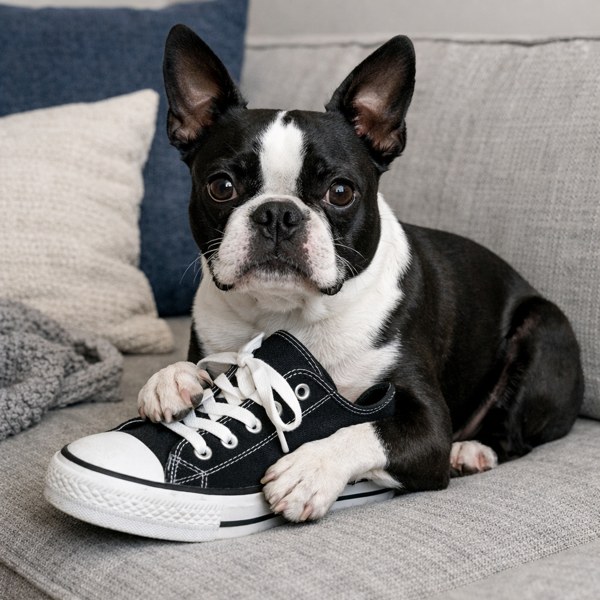}}{boston Terrier sitting on a couch with a shoe, and not a single sandwich}{sandwich}{GPT-Image-1.5} \\
\addlinespace
\textbf{2. Mental-state} & \excell{\includegraphics[width=\linewidth]{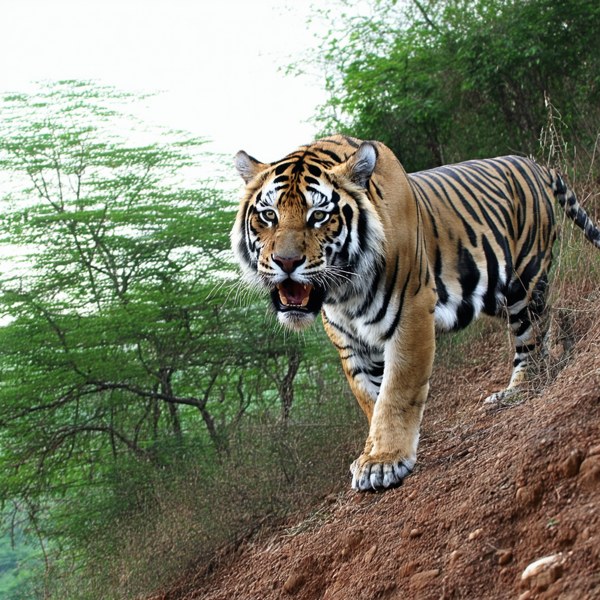}}{Maya at the ridge, believing that there is a fierce tiger at the ridge and being super afraid.}{tiger}{SD3.5-Large} & \excell{\includegraphics[width=\linewidth]{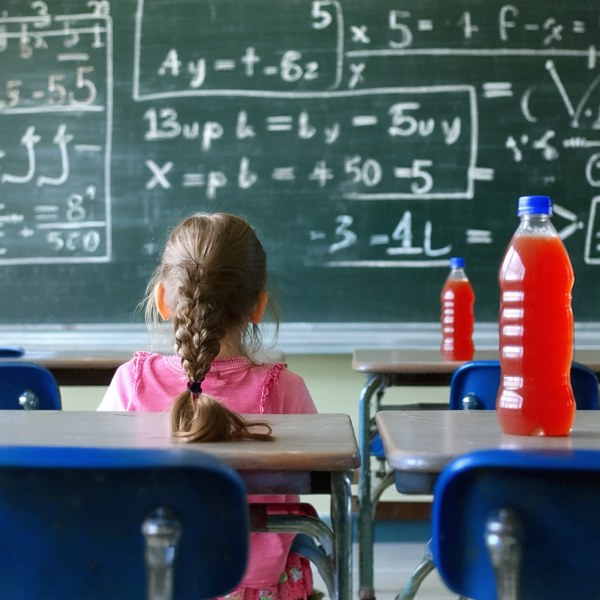}}{Sitting in math class, the thirsty girl stares at the whiteboard, wondering if she can buy a bottle of apple juice as soon as class ends.}{apple juice}{SD3.5-Large} & \excell{\includegraphics[width=\linewidth]{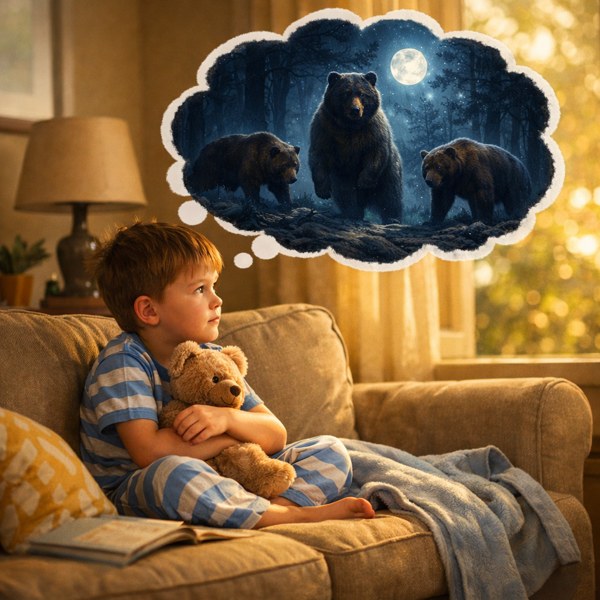}}{Sitting on the sofa one warm morning, the kid thought that there were bears outside at night, although he had never seen them.}{bears}{GPT-Image-1.5} & \excell{\includegraphics[width=\linewidth]{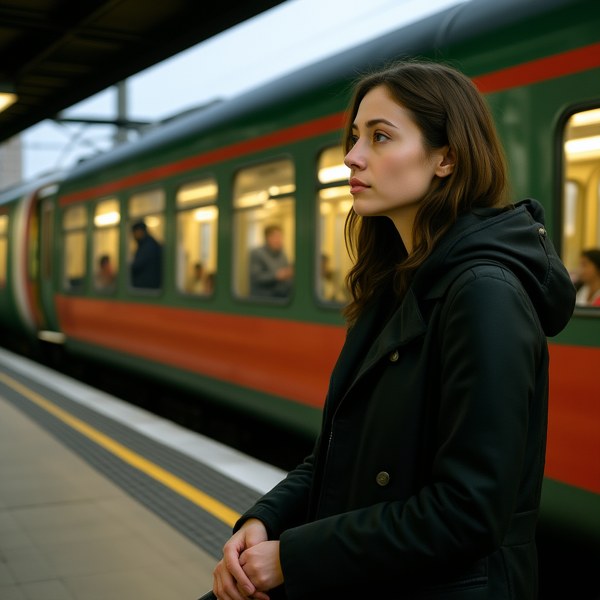}}{A woman waits at a train station, imagining her deceased friend arriving on the next train.}{friend}{OmniGen2} \\
\addlinespace
\textbf{3. Figurative} & \excell{\includegraphics[width=\linewidth]{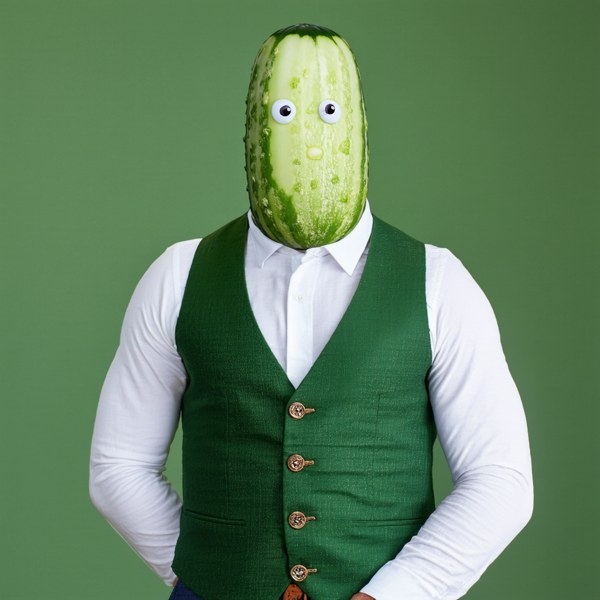}}{His vest is as green as a cucumber.}{cucumber}{SD3.5-Large} & \excell{\includegraphics[width=\linewidth]{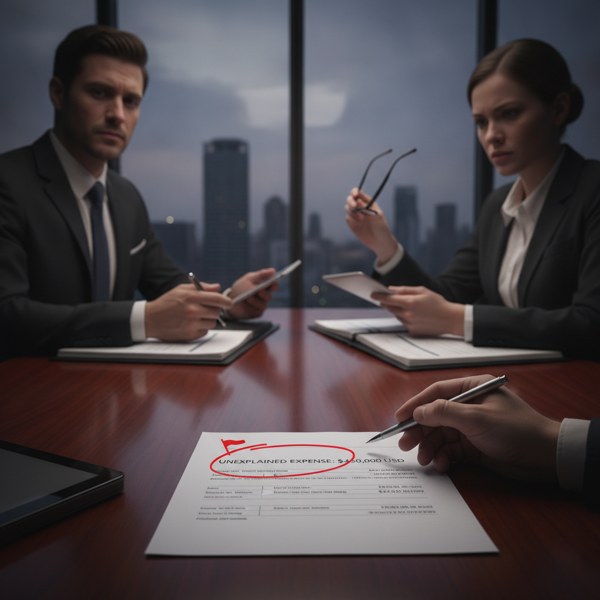}}{The unexplained expense was a red flag during an otherwise routine audit meeting.}{red flag}{Nano Banana} & \excell{\includegraphics[width=\linewidth]{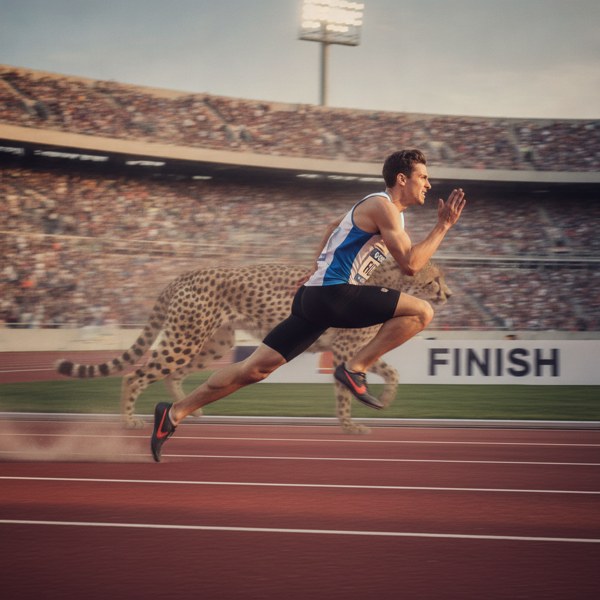}}{The runner is as fast as a cheetah as he races toward the finish line.}{cheetah}{Nano Banana} & \excell{\includegraphics[width=\linewidth]{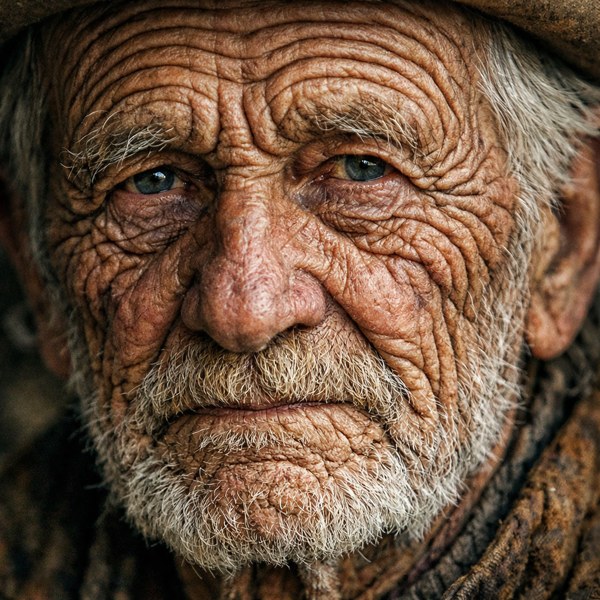}}{His wrinkles were as layered as tree rings.}{tree rings}{GPT-Image-1.5} \\
\addlinespace
\textbf{4. Perspectival} & \excell{\includegraphics[width=\linewidth]{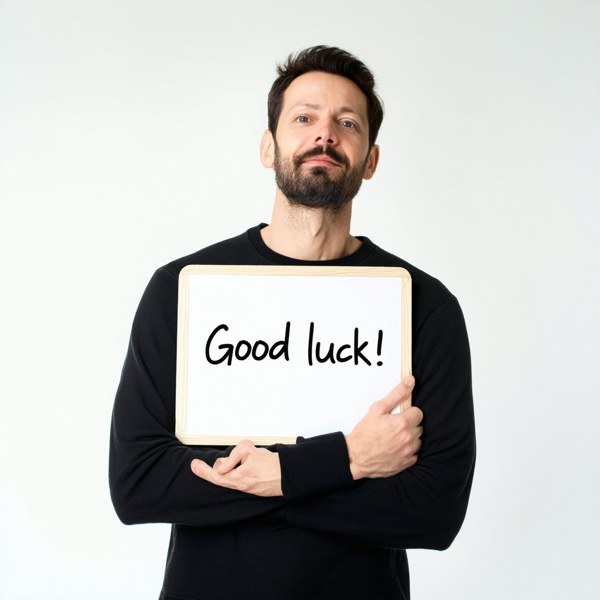}}{A man holds a whiteboard against his chest with ``Good luck!'' written on the hidden side}{Good luck}{SD3.5-Large} & \excell{\includegraphics[width=\linewidth]{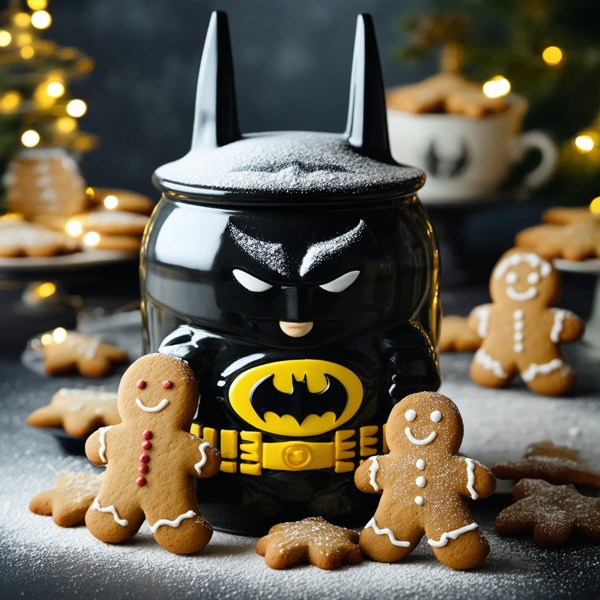}}{A closed batman cookie jar contains a gingerbread men}{gingerbread men}{SD3.5-Large} & \excell{\includegraphics[width=\linewidth]{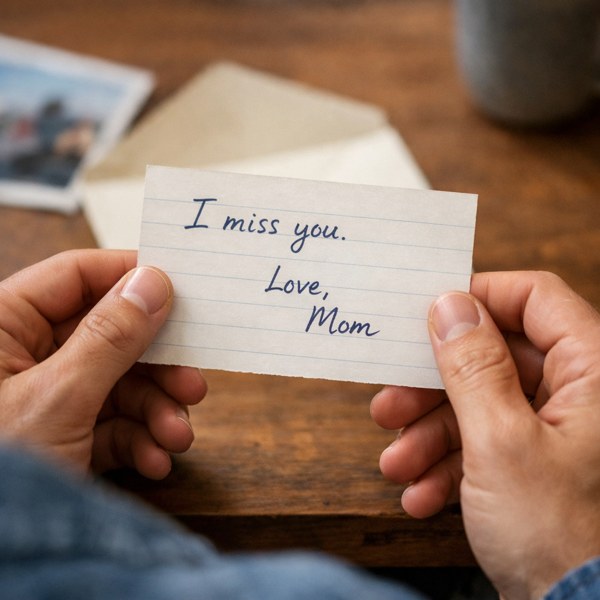}}{Tom is reading a letter. The letter is from his mom, which says: I miss you.}{I miss you}{GPT-Image-1.5} & \excell{\includegraphics[width=\linewidth]{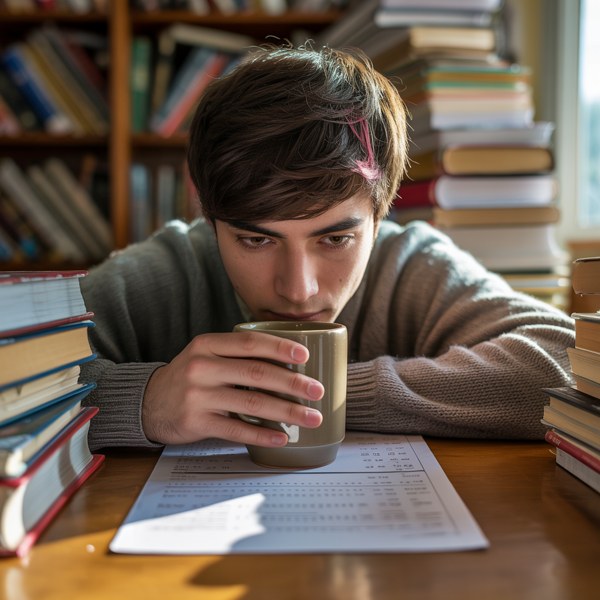}}{The student covered the grade on his exam sheet with a mug, pretending it wasn't a bad score.}{score}{Ideogram v3} \\
\bottomrule
\end{tabular}}
\caption{Taxonomy examples.}
\label{tab:taxonomy}
\end{table}

\subsection{Borderline cases}
\label{app:borderline}

\begin{figure}[htbp]
\centering
\newcommand{\bcell}[5]{%
  \begin{minipage}[t]{0.32\textwidth}\centering
    #1\\[2pt]
    {\scriptsize\raggedright prompt: \emph{#2}\\ target: \textbf{#3}\\ labels: #5\par}
  \end{minipage}}
\bcell{\includegraphics[width=\linewidth]{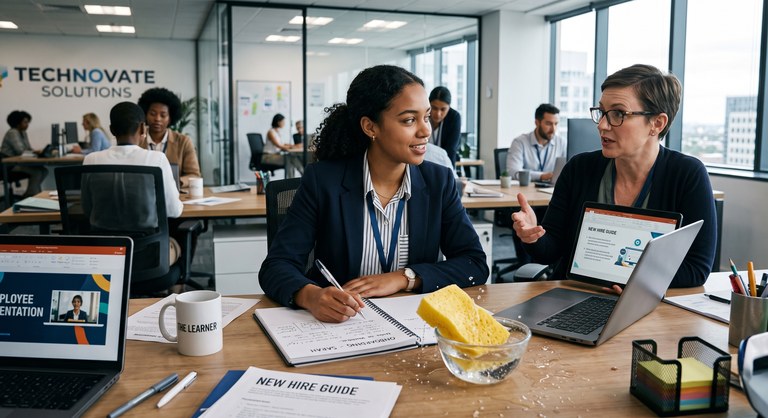}}{The new employee absorbed every instruction like a sponge soaking up water.}{sponge}{Gemini}{disruptive / silent}\hfill
\bcell{\includegraphics[width=\linewidth]{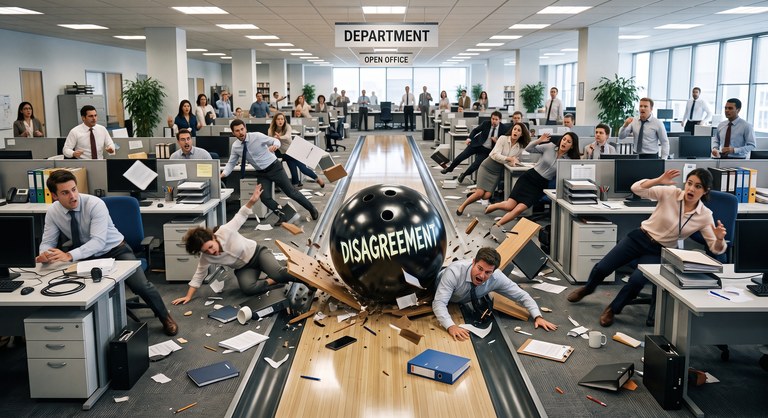}}{The tiny disagreement rolled through the department like a bowling ball knocking down everything in its path.}{bowling ball}{Gemini}{disruptive / integrated}\hfill
\bcell{\includegraphics[width=\linewidth]{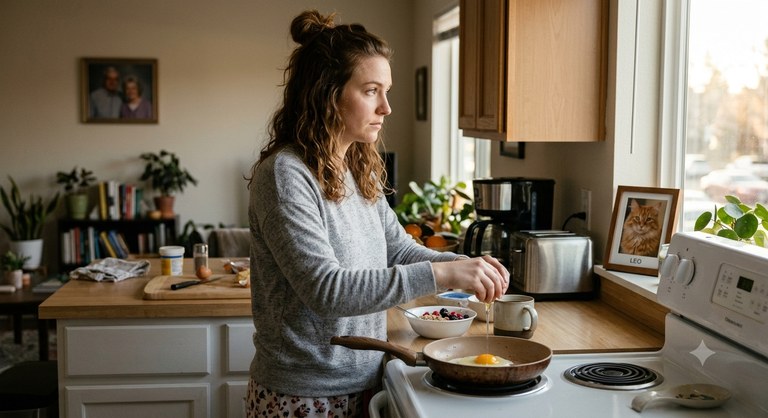}}{The woman is making her breakfast, wondering what her cat is doing, since it is living with her parents right now.}{cat}{Gemini}{silent / integrated}
\caption{Borderline cases: (a) disruptive or silent,
(b) disruptive or integrated, (c) integrated or silent. The two labels
under each image are the two annotators' labels.}
\label{fig:borderline}
\end{figure}

The outcome labels are defined by the effect that the target's appearance
has on the image, and for some images this effect is a matter of degree.
Figure~\ref{fig:borderline} shows three pilot images on which the two
annotators disagreed, one for each pair of labels that is hard to
separate.

\paragraph{Disruptive or silent.} The two labels differ on whether the
target's appearance fits the scene. In Figure~\ref{fig:borderline}(a), the sponge, the
vehicle of the comparison, is drawn into the office: one annotator read it as an object an office may plausibly hold, the other as
something that does not belong there.

\paragraph{Disruptive or integrated.} The two labels differ on whether an
exaggerated rendering is read as a stylized expression or as a scene that
breaks common sense. In Figure~\ref{fig:borderline}(b) a bowling ball rolls through the department:
one annotator took it as a stylization of the comparison, the other as an
object that a real office cannot contain.

\paragraph{Integrated or silent.} The two labels differ on whether a
depiction of the target inside the scene is a form the prompt allows or the
target slipping in. In Figure~\ref{fig:borderline}(c) the cat
appears as a framed photograph on the counter: one annotator read the
photograph as a fitting way to show an absent cat, the other as a cat
that was not supposed to appear at all.

\subsection{Unexpected over-realized targets}
\label{app:unexpected}

\begin{figure}[htbp]
\centering
\newcommand{\ucell}[4]{%
  \begin{minipage}[t]{0.48\textwidth}\centering
    #1\\[2pt]
    {\scriptsize\raggedright prompt: \emph{#2}\\ targets: #3\\ labels: #4\par}
  \end{minipage}}
\ucell{\includegraphics[width=\linewidth]{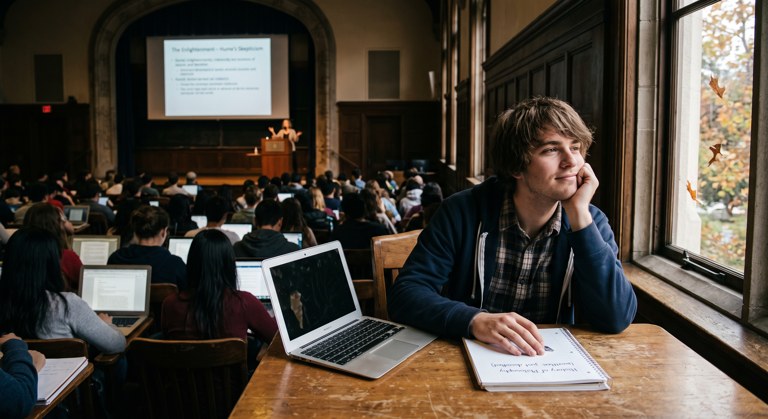}}{The student's attention drifted like a leaf in the wind during the long afternoon lecture.}{\textbf{leaf} (figurative), \textbf{the student} (perspectival)}{silent / disruptive}\hfill
\ucell{\includegraphics[width=\linewidth]{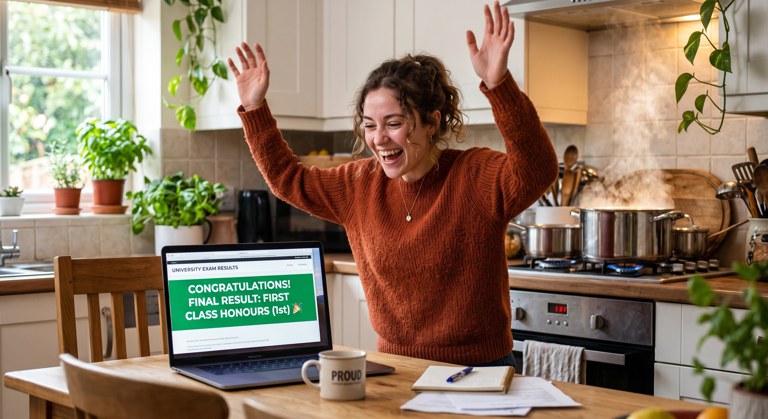}}{Her excitement bubbled over like a pot on the stove when the results came in.}{\textbf{pot} (figurative), \textbf{the results} (perspectival)}{silent / disruptive}
\caption{Two pilot images (Gemini) that over-realize two targets each: the
vehicle of the comparison, silently, and the viewpoint of the person,
disruptively. (a) The student faces the camera instead of the lecturer.
(b) The screen faces the camera instead of the woman reading it.}
\label{fig:unexpected}
\end{figure}

An annotator writes a
prompt with one target in mind, but the generator may over-realize another
mention, and even another family. Figure~\ref{fig:unexpected} shows two
example images which over-realize
both the vehicle of the comparison and the viewpoint. In (a) the leaves
drift past the window (silent), and the student, who should face the
lecturer, is turned toward the camera so that his face is visible
(disruptive). In (b) a pot bubbles on the stove (silent), and the screen
that carries the results faces the camera rather than the woman who is
reading it (disruptive). 

In practical use, a generator is not told which mention to withhold, and a
single prompt can over-realize several of them. Perspectival
over-realization in particular can arise in any prompt that places a person
or a text-bearing object in a scene, without a cue word. Methods that suppress a given word, such as a negative prompt or a
reweighting of the attention on that word, must first be told which word to
suppress, and a user does not say.

\section{Dataset Details}
\label{app:dataset}

\subsection{Statistics}
\label{app:datastats}

Table~\ref{tab:datastats} gives the size of the raw and
expanded splits of OverReal-Gen, the size of the gold and ambiguous splits
of OverReal-Det, the label distribution of the gold split, and the
agreement between the two annotators. \rev{When writing the three hand-written
families, the annotators were asked not to reuse one template, so that the
prompts cover diverse failures.}

\begin{table}[h]
\centering
\small
\setlength{\tabcolsep}{3.5pt}
\resizebox{\textwidth}{!}{%
\begin{tabular}{l cc cc c cccc cc}
\toprule
\multirow{3}{*}{Family} & \multicolumn{4}{c}{\textbf{OverReal-Gen}} & \multicolumn{7}{c}{\textbf{OverReal-Det}} \\
\cmidrule(lr){2-5} \cmidrule(lr){6-12}
& \multicolumn{2}{c}{Raw} & \multicolumn{2}{c}{Expanded} & \multirow{2}{*}{\shortstack{Gold\\size}} & \multicolumn{4}{c}{Label distribution (\%)} & \multirow{2}{*}{\shortstack{Amb.\\size}} & \multirow{2}{*}{\shortstack{Overall\\agreement ($\kappa$)}} \\
\cmidrule(lr){2-3} \cmidrule(lr){4-5} \cmidrule(lr){7-10}
& Size & Avg.\ tokens & Size & Avg.\ tokens & & DO & SO & I & W & & \\
\midrule
Existence-canceling & 200 & 16.9 & 400 & 112.1 & 204 & 32.4 & 0.5 & 0.0 & 67.2 & 23  & 0.73 \\
Mental-state        & 75  & 23.5 & 150 & 124.0 & 239 & 53.6 & 18.0 & 11.3 & 17.2 & 112 & 0.51 \\
Figurative          & 237 & 16.4 & 474 & 116.4 & 932 & 36.6 & 28.1 & 2.9 & 32.4 & 299 & 0.65 \\
Perspectival        & 186 & 23.1 & 372 & 118.7 & 878 & 70.5 & 10.3 & 7.3 & 12.0 & 154 & 0.72 \\
\midrule
Overall & 698 & 19.1 & 1{,}396 & 117.8 & 2{,}253 & 51.2 & 17.6 & 5.2 & 26.0 & 588 & 0.68 \\
\bottomrule
\end{tabular}}
\caption{OverReal statistics. Label distribution: share of the gold split labeled
disruptive over-realization (DO), silent over-realization (SO), integrated realization (I) and withheld realization (W).}
\label{tab:datastats}
\end{table}

\subsection{The ambiguous split}
\label{app:ambiguous}

An image enters the ambiguous split when the label sets of its two
annotators are disjoint (\S\ref{sec:overreal-det}). Table~\ref{tab:ambiguous}
sorts the 588 such images, 20.7\% of the 2{,}841 annotated images,
by what the two annotators disagree on. In 245 of them, the annotators
agree on whether the target is over-realized;
in 266 of them, one annotator read the image
as over-realized and the other as correct; the remaining 77 involve the label
\emph{other}.

\begin{table}[h]
\centering
\small
\begin{tabular}{l cccc c}
\toprule
& Existence-canceling & Mental-state & Figurative & Perspectival & All \\
\midrule
Form of the outcome only        & 6  & 52 & 125 & 62 & 245 \\
Over-realized against correct   & 13 & 43 & 141 & 69 & 266 \\
One label is \emph{other}       & 4  & 17 & 33  & 23 & 77 \\
\bottomrule
\end{tabular}
\caption{The ambiguous split of OverReal-Det by the kind of disagreement.}
\label{tab:ambiguous}
\end{table}

\paragraph{\rev{Stability of the rates under the ambiguous images.}}
\rev{The headline rates (Figure~\ref{fig:main}, Table~\ref{tab:mainresults})
come from the automatic annotator, so the ambiguous images do not enter
them directly. To see how much they could matter, we recompute the
over-realization rate on OverReal-Det with the ambiguous images included,
taking each annotator's labels in turn instead of the gold labels. The
rate per family moves by at most $0.04$ (existence-canceling $0.33$ to
$0.36$ and $0.37$, mental-state $0.71$ to $0.73$ and $0.72$, figurative
$0.65$ to $0.65$ and $0.64$, perspectival $0.81$ to $0.79$ and $0.79$),
and the generator rankings stay almost unchanged (Spearman
$\rho \geq 0.77$ in every family).}

\section{Automatic Annotator}
\label{app:autoannot-val}

\subsection{Cascade design}
\label{app:cascade}

\begin{figure}[htbp]
\centering
\includegraphics[width=0.95\textwidth]{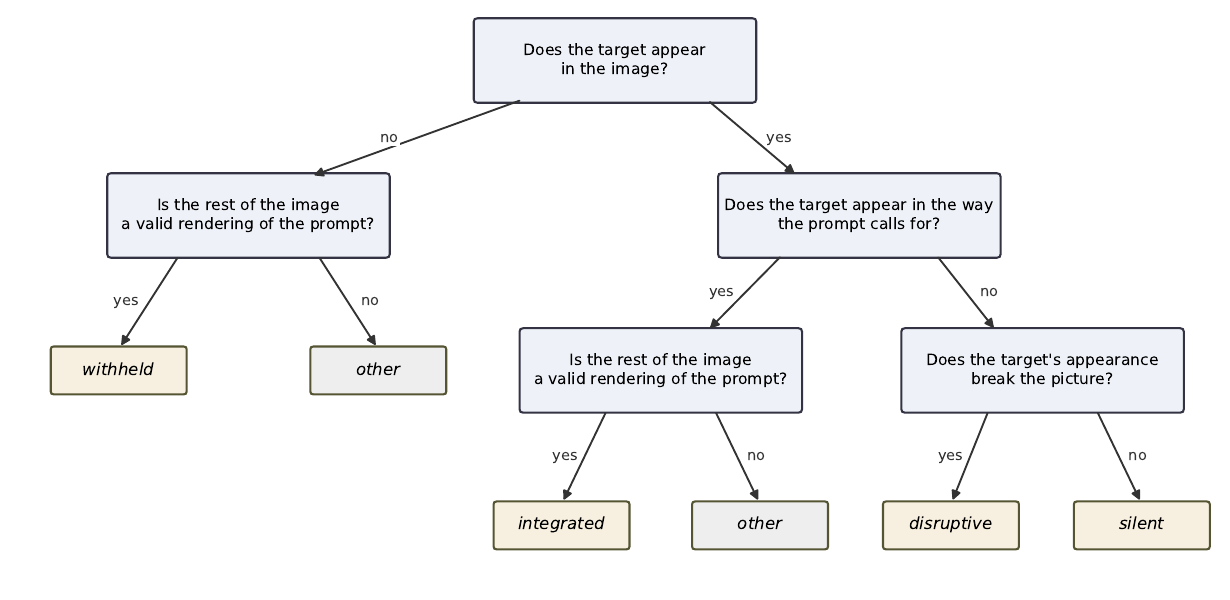}
\caption{The annotation cascade as a decision tree. Each decision node
(blue) is one yes/no question put to the annotator in a separate call, with
the image and the prompt; the leaves (italic) are the labels. The family-specific wording of the questions is given
in Appendix~\ref{app:autoannot}.}
\label{fig:cascade}
\end{figure}

Rather than asking for one of four labels at once, the annotator follows
the decision tree of Figure~\ref{fig:cascade}, one yes/no question per
call. The first question is whether the target appears in the image. If it
does not, the image is \emph{withheld} when the rest of the image is a
valid rendering of the prompt, and \emph{other} when it is not. If it does,
the next question is whether the target appears in the way the prompt calls
for. If so, the same validity question decides between \emph{integrated}
and \emph{other}; if not, the image is
\emph{disruptive} when the target's appearance breaks the picture, and
\emph{silent} when it does not. The questions that instantiate these
decisions differ by family for higher precision. The full question set per family is given in
Appendix~\ref{app:autoannot}.

\subsection{Self-evolution of the AI annotator}
\label{app:autoannot-dev}

The AI annotator is adapted to the task without updating any model weights,
by a form of self-evolution: the model annotates a development set with its
current questions, revises the questions from the mistakes it made, and
repeats until its agreement with the human labels stops improving. We build
it on the human labels of OverReal-Det, split evenly, stratified by family
and generator, into a development split and a test split; the development
split alone drives the self-evolution, and the frozen annotator is then evaluated
once on the test set. In each round the model scores the development split,
receives the images on which its label disagrees with the human one, traces
each disagreement to the question that produced it, and proposes a rewrite
of that question; we accept the rewrite and score the revised questions
again. The best wording is selected per family.

\subsection{Agreement with the human labels}
\label{app:autoannot-results}

The annotator uses Claude Opus~5 \citep{anthropic2026opus5} as the base model.
Table~\ref{tab:autoannot} reports the AI-human agreement and human-human agreement in Cohen's $\kappa$. Conclusions are given as
follows.

\paragraph{Self-evolution improves the cascade annotator.} Asking the model for
one of the four labels outright is unreliable: a multiple-choice prompt
over the four labels reached only $\kappa=0.46$ against the human labels.
Self-evolution of the cascade questions then raised the agreement with the
human labels on the development split from $\kappa=0.61$ to $0.68$ over
the four labels, and from $0.65$ to $0.76$ on the split between over-realization and
correct output.

\paragraph{The annotator agrees with the human labels nearly as well as
the humans agree with each other.} Over the four labels, $\kappa$ is
$0.58$ overall against $0.70$ between the humans; on the split between
over-realization and correct output, which the rates of
Table~\ref{tab:mainresults} rest on, it is $0.70$ against $0.75$, a
substantial agreement. The automatic labels are therefore reliable enough
for the rates we report, if not for a fine reading of every image.

\paragraph{\rev{The agreement holds on prompts the annotator never saw during
development.}} \rev{The development and test splits of OverReal-Det were drawn
over images, so 87\% of the 1{,}118 test images share a prompt with a
development image. On the 148 test images whose prompt never appears in
the development split, $\kappa$ is $0.54$ over the four labels and $0.66$
on the split between over-realization and correct output, against $0.58$
and $0.70$ on the full test split.}

\paragraph{Part of the gap to the humans is the ambiguity of the task.}
The $\kappa$ values in Table~\ref{tab:autoannot} are not uniformly high,
but they come close to the agreement between the humans on this task.
Deciding whether the target appears is easy for both the annotator and the
humans ($\kappa=0.83$ against $0.86$). Deciding what its appearance amounts
to is harder for both, because the outcome labels are defined by the effect
that the target's appearance has on the image, and this effect is a matter
of degree, as described in the limitations; on mental-state prompts the
annotator is close to the humans ($0.42$ against $0.47$). On perspectival
prompts it falls short of them ($0.41$ against $0.75$), so there the gap is
the model's. \rev{The label with the least sharp boundary is \emph{integrated}
(Appendix~\ref{app:borderline}): on whether an image is integrated, the two
humans reach $\kappa=0.52$.}

\begin{table}[h]
\centering
\small
\setlength{\tabcolsep}{4pt}
\resizebox{\textwidth}{!}{%
\begin{tabular}{ll ccccc}
\toprule
Decision & & Existence-canceling & Mental-state & Figurative & Perspectival & Overall \\
\midrule
\multirow{2}{*}{Four labels} & AI-human & 1.00 & 0.42 & 0.60 & 0.41 & 0.58 \\
 & humans & 0.83 & 0.47 & 0.67 & 0.75 & 0.70 \\
\midrule
\multirow{2}{*}{Target appears} & AI-human & 1.00 & 0.74 & 0.82 & 0.71 & 0.83 \\
 & humans & 0.88 & 0.84 & 0.83 & 0.87 & 0.86 \\
\midrule
\multirow{2}{*}{Over-realized vs.\ correct} & AI-human & 1.00 & 0.53 & 0.69 & 0.60 & 0.70 \\
 & humans & 0.88 & 0.49 & 0.74 & 0.79 & 0.75 \\
\bottomrule
\end{tabular}}
\caption{Cohen's $\kappa$ of the automatic annotator (Claude Opus~5) against
the human gold labels on the held-out test split of OverReal-Det (1{,}118
images), and of the two human annotators against each other on the 2{,}688
doubly annotated images with one label each, for the four-outcome label,
for whether the target appears, and for the collapse to over-realized
against correct.}
\label{tab:autoannot}
\end{table}

\section{Results}
\label{app:results}
\subsection{Conventional Evaluation: LLM Judge Details}
\label{app:judges}

\begin{table}[t]
\centering
\scriptsize
\setlength{\tabcolsep}{1pt}
\begin{tabular}{l ccccc@{\hspace{7pt}}ccccc@{\hspace{7pt}}ccccc@{\hspace{7pt}}ccccc}
\toprule
& \multicolumn{5}{c}{\textbf{Existence-canceling}} & \multicolumn{5}{c}{\textbf{Mental-state}} & \multicolumn{5}{c}{\textbf{Figurative}} & \multicolumn{5}{c}{\textbf{Perspectival}} \\
\cmidrule(lr){2-6} \cmidrule(lr){7-11} \cmidrule(lr){12-16} \cmidrule(lr){17-21}
& CS{\tiny$\uparrow$} & VQ{\tiny$\uparrow$} & PS{\tiny$\uparrow$} & HP{\tiny$\uparrow$} & LJ{\tiny$\uparrow$} & CS{\tiny$\uparrow$} & VQ{\tiny$\uparrow$} & PS{\tiny$\uparrow$} & HP{\tiny$\uparrow$} & LJ{\tiny$\uparrow$} & CS{\tiny$\uparrow$} & VQ{\tiny$\uparrow$} & PS{\tiny$\uparrow$} & HP{\tiny$\uparrow$} & LJ{\tiny$\uparrow$} & CS{\tiny$\uparrow$} & VQ{\tiny$\uparrow$} & PS{\tiny$\uparrow$} & HP{\tiny$\uparrow$} & LJ{\tiny$\uparrow$} \\
\midrule
GPT-Image    & .67 & \textbf{.87} & \textbf{23.3} & \textbf{.31} & \textbf{9.4} & \textbf{.75} & .78 & \textbf{22.1} & \textbf{.33} & \textbf{8.2} & \textbf{.69} & \textbf{.84} & \textbf{22.1} & \textbf{.31} & \textbf{7.5} & .74 & .66 & 22.4 & .29 & \textbf{5.6} \\
Nano Banana  & .64 & .82 & 22.9 & .28 & 8.7 & .71 & \textbf{.78} & 21.8 & .30 & 7.8 & .63 & .83 & 21.6 & .28 & \textbf{7.5} & .71 & .76 & 22.3 & .28 & 4.5 \\
Ideogram     & .61 & .82 & 22.7 & .29 & 8.3 & .66 & .62 & 21.2 & .29 & 6.9 & .57 & .73 & 21.1 & .27 & 6.9 & .70 & .70 & 22.4 & .30 & 3.6 \\
\midrule
FLUX.1-dev   & .66 & .59 & 22.9 & .30 & 5.8 & .73 & .66 & 21.7 & .30 & 6.2 & .62 & .70 & 21.4 & .29 & 5.5 & .73 & .69 & 22.6 & .30 & 3.2 \\
Qwen-Image   & \textbf{.73} & .51 & 23.2 & .31 & 5.2 & .75 & .75 & 22.0 & .32 & 6.7 & .66 & .79 & 21.7 & .30 & 6.3 & \textbf{.77} & \textbf{.77} & \textbf{22.8} & \textbf{.31} & 3.3 \\
OmniGen2     & .70 & .58 & 22.9 & .30 & 5.6 & .71 & .64 & 21.5 & .32 & 5.8 & .63 & .71 & 21.2 & .28 & 5.3 & .75 & .72 & 22.3 & .30 & 3.0 \\
SD3.5-Medium & .69 & .69 & 22.5 & .28 & 6.5 & .72 & .65 & 21.4 & .29 & 6.0 & .62 & .69 & 21.1 & .26 & 5.5 & .76 & .69 & 21.9 & .28 & 3.2 \\
SD3.5-Large  & .71 & .50 & 22.9 & .30 & 5.0 & .73 & .67 & 21.8 & .31 & 6.3 & .66 & .71 & 21.6 & .29 & 5.4 & .75 & .71 & 22.5 & .30 & 3.6 \\
\bottomrule
\end{tabular}
\caption{Conventional evaluation of the eight systems on the raw prompts of OverReal-Gen: mean CLIPScore (CS), VQAScore (VQ), PickScore (PS), HPSv2 (HP) and LLM judge (LJ) per family.}
\label{tab:conventional}
\end{table}

\begin{table}[h]
\centering
\scriptsize
\setlength{\tabcolsep}{1pt}
\begin{tabular}{l ccccc@{\hspace{7pt}}ccccc@{\hspace{7pt}}ccccc@{\hspace{7pt}}ccccc}
\toprule
& \multicolumn{5}{c}{\textbf{Existence-canceling}} & \multicolumn{5}{c}{\textbf{Mental-state}} & \multicolumn{5}{c}{\textbf{Figurative}} & \multicolumn{5}{c}{\textbf{Perspectival}} \\
\cmidrule(lr){2-6} \cmidrule(lr){7-11} \cmidrule(lr){12-16} \cmidrule(lr){17-21}
Generator & DO{\tiny$\downarrow$} & SO{\tiny$\downarrow$} & I{\tiny$\uparrow$} & W{\tiny$\uparrow$} & $u${\tiny$\downarrow$} & DO{\tiny$\downarrow$} & SO{\tiny$\downarrow$} & I{\tiny$\uparrow$} & W{\tiny$\uparrow$} & $u${\tiny$\downarrow$} & DO{\tiny$\downarrow$} & SO{\tiny$\downarrow$} & I{\tiny$\uparrow$} & W{\tiny$\uparrow$} & $u${\tiny$\downarrow$} & DO{\tiny$\downarrow$} & SO{\tiny$\downarrow$} & I{\tiny$\uparrow$} & W{\tiny$\uparrow$} & $u${\tiny$\downarrow$} \\
\midrule
\multicolumn{21}{l}{\emph{(a) OverReal-Gen}} \\
GPT-Image & \textbf{.02} & .00 & .00 & \textbf{.98} & .02 & .35 & .26 & .28 & .11 & .08 & .22 & .29 & \textbf{.16} & .33 & .05 & \textbf{.54} & .15 & \textbf{.03} & \textbf{.28} & .05 \\
Nano Banana & .12 & .00 & .00 & .88 & .03 & .37 & \textbf{.19} & \textbf{.37} & .07 & .09 & .15 & .26 & .15 & .43 & .09 & .78 & .17 & .02 & .03 & .07 \\
Ideogram & .08 & .00 & .00 & .92 & .02 & \textbf{.29} & .24 & .03 & \textbf{.44} & \textbf{.02} & \textbf{.07} & \textbf{.25} & .03 & \textbf{.65} & .04 & .65 & .18 & .00 & .17 & \textbf{.02} \\
\midrule
FLUX.1-dev & .43 & .00 & .00 & .57 & .02 & .44 & .28 & .03 & .24 & \textbf{.02} & .09 & .30 & .14 & .46 & \textbf{.02} & .56 & .19 & .01 & .24 & .03 \\
Qwen-Image & .64 & .00 & .00 & .36 & .01 & .67 & .22 & .01 & .10 & .03 & .28 & .33 & .07 & .31 & \textbf{.02} & .67 & .22 & .01 & .10 & .04 \\
OmniGen2 & .51 & .00 & .00 & .49 & .01 & .64 & .22 & .03 & .11 & \textbf{.02} & .19 & .27 & .08 & .46 & .03 & .64 & .16 & .01 & .18 & .03 \\
SD3.5-Medium & .34 & .00 & .00 & .66 & \textbf{.00} & .66 & .21 & .01 & .12 & \textbf{.02} & .09 & .31 & .06 & .54 & \textbf{.02} & .67 & \textbf{.11} & .00 & .22 & \textbf{.02} \\
SD3.5-Large & .64 & .00 & .00 & .36 & .01 & .61 & .23 & .03 & .13 & \textbf{.02} & .20 & .40 & .11 & .28 & \textbf{.02} & .59 & .16 & .02 & .22 & .03 \\
\midrule
\multicolumn{21}{l}{\emph{(b) In-the-wild prompts}} \\
GPT-Image    & .28 & .00 & .00 & .72 & --- & .21 & .12 & .64 & .03 & --- & .01 & .21 & .47 & .31 & --- & .59 & .23 & .00 & .18 & --- \\
Nano Banana  & .38 & .00 & .00 & .62 & --- & .26 & .10 & .58 & .06 & --- & .01 & .24 & .35 & .40 & --- & .70 & .07 & .01 & .21 & --- \\
Ideogram     & .55 & .00 & .00 & .45 & --- & .28 & .16 & .14 & .41 & --- & .02 & .27 & .26 & .45 & --- & .74 & .15 & .00 & .12 & --- \\
\midrule
FLUX.1-dev   & .70 & .00 & .00 & .30 & --- & .30 & .17 & .04 & .49 & --- & .02 & .29 & .14 & .54 & --- & .65 & .13 & .00 & .22 & --- \\
Qwen-Image   & .72 & .00 & .00 & .28 & --- & .51 & .24 & .06 & .19 & --- & .04 & .26 & .36 & .34 & --- & .75 & .12 & .00 & .13 & --- \\
OmniGen2     & .67 & .00 & .00 & .33 & --- & .42 & .18 & .04 & .36 & --- & .02 & .26 & .26 & .46 & --- & .66 & .09 & .00 & .25 & --- \\
SD3.5-Medium & .69 & .00 & .00 & .31 & --- & .33 & .20 & .03 & .44 & --- & .03 & .26 & .23 & .47 & --- & .63 & .10 & .00 & .26 & --- \\
SD3.5-Large  & .70 & .00 & .00 & .30 & --- & .38 & .21 & .03 & .38 & --- & .02 & .26 & .29 & .43 & --- & .69 & .10 & .01 & .19 & --- \\
\bottomrule
\end{tabular}

\caption{Main results: proportion of images under each outcome label
(definition in \S\ref{sec:spectrum}), namely disruptive over-realization
(DO), silent over-realization (SO), integrated realization (I) and withheld
realization (W), and the mean unprompted appearance rate ($u$, definition in
\S\ref{sec:autoannot}). (a) On the curated prompts of OverReal-Gen. (b) On
the 397 in-the-wild prompts of \S\ref{sec:main}, one image per prompt and
system; $u$ was not measured there.}
\label{tab:mainresults}
\end{table}

Both judges score the same images with the verbatim VIEScore
text-to-image semantic-consistency template: Claude Opus~5
(\texttt{claude-opus-5}, accessed 2026-09) and GPT-5.2
(\texttt{gpt-5.2-2025-12-11}). Table~\ref{tab:conventional} reports the Opus
rating; Table~\ref{tab:ljgpt} gives the GPT-5.2 rating for the same cells and
Table~\ref{tab:judgeagreement} the agreement of the two judges.

\begin{table}[h]
\centering
\small
\begin{tabular}{l cccc}
\toprule
& Existence-canceling & Mental-state & Figurative & Perspectival \\
\midrule
GPT-Image    & 9.5 & 8.8 & 9.2 & 6.3 \\
Nano Banana  & 9.1 & 8.8 & 9.0 & 6.2 \\
Ideogram     & 9.0 & 7.8 & 8.2 & 5.2 \\
\midrule
FLUX.1-dev   & 6.2 & 7.7 & 7.9 & 4.9 \\
Qwen-Image   & 5.0 & 8.3 & 8.7 & 5.3 \\
OmniGen2     & 6.0 & 7.6 & 8.0 & 5.0 \\
SD3.5-Medium & 7.1 & 7.4 & 7.4 & 5.0 \\
SD3.5-Large  & 4.9 & 7.6 & 8.1 & 5.0 \\
\bottomrule
\end{tabular}
\caption{LLM judge (LJ) ratings by GPT-5.2, the same cells as the LJ column
of Table~\ref{tab:conventional}.}
\label{tab:ljgpt}
\end{table}

\begin{table}[h]
\centering
\small
\begin{tabular}{l cc cc c}
\toprule
Family & Opus mean & GPT mean & Pearson $r$ & Mean $|\Delta|$ & $n$ \\
\midrule
Existence-canceling & 6.86 $\pm$ 2.83 & 7.37 $\pm$ 3.27 & 0.90 & 1.02 & 1700 \\
Mental-state & 6.78 $\pm$ 1.52 & 7.97 $\pm$ 1.30 & 0.62 & 1.32 & 1750 \\
Figurative & 6.71 $\pm$ 1.95 & 8.29 $\pm$ 1.39 & 0.57 & 1.69 & 1800 \\
Perspectival & 3.63 $\pm$ 1.81 & 5.27 $\pm$ 2.38 & 0.56 & 1.98 & 1800 \\
\midrule
Overall & 5.98 $\pm$ 2.49 & 7.22 $\pm$ 2.52 & 0.78 & 1.51 & 7050 \\
\bottomrule
\end{tabular}
\caption{Between-judge statistics: mean rating $\pm$ standard deviation per
judge, correlation, and mean absolute rating difference.}
\label{tab:judgeagreement}
\end{table}

\subsection{Intervals for the over-realization rates}
\label{app:ci}

Table~\ref{tab:ci} gives the over-realization rate of Table~\ref{tab:mainresults},
the sum of the DO and SO columns, with half the width of a 95\% bootstrap
interval. 

\begin{table}[t]
\centering
\small
\setlength{\tabcolsep}{6pt}
\begin{tabular}{l cccc c}
\toprule
& Existence-canceling & Mental-state & Figurative & Perspectival & All families \\
\midrule
GPT-Image     & .02 $\pm$ .03 & .61 $\pm$ .12 & .51 $\pm$ .11 & .69 $\pm$ .10 & .43 $\pm$ .05 \\
Nano Banana   & .12 $\pm$ .09 & .56 $\pm$ .11 & .41 $\pm$ .10 & .95 $\pm$ .04 & .49 $\pm$ .05 \\
Ideogram      & .08 $\pm$ .07 & .53 $\pm$ .11 & .32 $\pm$ .09 & .83 $\pm$ .08 & .41 $\pm$ .04 \\
\midrule
FLUX.1-dev    & .43 $\pm$ .13 & .73 $\pm$ .11 & .39 $\pm$ .10 & .75 $\pm$ .10 & .53 $\pm$ .06 \\
Qwen-Image    & .64 $\pm$ .13 & .89 $\pm$ .07 & .62 $\pm$ .10 & .89 $\pm$ .06 & .73 $\pm$ .05 \\
OmniGen2      & .51 $\pm$ .11 & .86 $\pm$ .08 & .45 $\pm$ .11 & .81 $\pm$ .08 & .61 $\pm$ .05 \\
SD3.5-Medium  & .34 $\pm$ .11 & .87 $\pm$ .07 & .40 $\pm$ .10 & .78 $\pm$ .09 & .54 $\pm$ .05 \\
SD3.5-Large   & .64 $\pm$ .12 & .84 $\pm$ .08 & .61 $\pm$ .10 & .76 $\pm$ .10 & .68 $\pm$ .06 \\
\bottomrule
\end{tabular}
\caption{Over-realization rate of each system per family and over all
families, weighting the families by their prompt counts in OverReal-Gen,
$\pm$ half the width of a 95\% bootstrap interval over prompts.}
\label{tab:ci}
\end{table}

\subsection{Outcomes across the two seeds}
\label{app:seeds}

For the raw prompts of OverReal-Gen, each model generates two images
per prompt under different seeds. Table~\ref{tab:seeds} conditions on the
label of one image and gives the distribution of the label of the other,
over the labeled sample of \S\ref{sec:autoannot} (172 silent, 428 correct
and 588 disruptive conditioning images). Existence-canceling prompts have
no silent images.

\begin{table}[t]
\centering
\small
\setlength{\tabcolsep}{4pt}
\begin{tabular}{l ccccc}
\toprule
& Existence-canceling & Mental-state & Figurative & Perspectival & All \\
\midrule
$P(\mathrm{DO}_1 \mid \mathrm{DO}_0)$ & .84  & .77  & .65  & .71  & .79 \\
$P(\mathrm{SO}_1 \mid \mathrm{DO}_0)$ & .00  & .17  & .10  & .20  & .10 \\
$P(\mathrm{C}_1 \mid \mathrm{DO}_0)$  & .16  & .06  & .26  & .09  & .12 \\
\midrule
$P(\mathrm{DO}_1 \mid \mathrm{SO}_0)$ & ---  & .44  & .05  & .64  & .33 \\
$P(\mathrm{SO}_1 \mid \mathrm{SO}_0)$ & ---  & .50  & .76  & .32  & .57 \\
$P(\mathrm{C}_1 \mid \mathrm{SO}_0)$  & ---  & .06  & .19  & .04  & .10 \\
\midrule
$P(\mathrm{DO}_1 \mid \mathrm{C}_0)$  & .16  & .23  & .09  & .19  & .16 \\
$P(\mathrm{SO}_1 \mid \mathrm{C}_0)$  & .00  & .09  & .13  & .03  & .04 \\
$P(\mathrm{C}_1 \mid \mathrm{C}_0)$   & .84  & .68  & .78  & .78  & .80 \\
\bottomrule
\end{tabular}
\caption{Label of the other seed's image given the label of one image of
the same prompt and open model, on the raw prompts of OverReal-Gen.
$P(y_1 \mid y_0)$ is the share of images with label $y_0$ whose other-seed
image has label $y_1$; DO disruptive, SO silent, C correct (integrated or
withheld). Existence-canceling prompts have no silent images.}
\label{tab:seeds}
\end{table}

\subsection{Conventional metrics on matched pairs}
\label{app:paired}

For the raw prompts of OverReal-Gen, each model generates two images per
prompt under different seeds. Over the labeled sample of
\S\ref{sec:autoannot}, we take the pairs whose two images differ in
outcome, one over-realized and one correct, and ask which of the two each
conventional metric scores higher. Table~\ref{tab:paired} gives this
winning rate of the over-realized image, by kind of over-realization and
per family.

\begin{table}[t]
\centering
\small
\setlength{\tabcolsep}{4pt}
\resizebox{\textwidth}{!}{%
\begin{tabular}{l c | cc | cccc}
\toprule
& & \multicolumn{2}{c|}{by kind of over-realization} & \multicolumn{4}{c}{by family} \\
\cmidrule(lr){3-4} \cmidrule(lr){5-8}
Metric & Winning rate & Disruptive & Silent & Existence-canceling & Mental-state & Figurative & Perspectival \\
\midrule
CLIPScore  & .64 $\pm$ .10 & .69 & .44 & .75 & .56 & .60 & .38 \\
VQAScore   & .33 $\pm$ .13 & .25 & .80 & .05 & .67 & .75 & .50 \\
PickScore  & .66 $\pm$ .10 & .72 & .44 & .72 & .67 & .60 & .50 \\
HPSv2      & .65 $\pm$ .11 & .69 & .50 & .65 & .72 & .70 & .38 \\
LLM judge  & .12 $\pm$ .08 & .06 & .50 & .00 & .27 & .33 & .17 \\
\bottomrule
\end{tabular}}
\caption{Winning rate: share of matched pairs (same prompt, same open
model, two seeds; one image over-realized, one correct) in which the
metric scores the over-realized image higher, $\pm$ half the width of a
95\% bootstrap interval over prompts.}
\label{tab:paired}
\end{table}

\paragraph{The conventional metrics reward over-realization or are indifferent to it.}
CLIPScore, PickScore and HPSv2 score the over-realized image higher in
64\% to 66\% of the pairs. VQAScore is split: on existence-canceling prompts it
scores the correct image higher in 95\% of the pairs, but on mental-state and figurative prompts it scores the
over-realized image higher in 67\% and 75\%, and on perspectival prompts it scores either image higher equally often (50\%). The LLM judge scores the
correct image higher in 88\%. The preference of the three metrics is
strongest on the disruptive images (69\% to 72\%); on the pairs whose
over-realized image is silent they score either image higher about equally
often (44\% to 50\%), so they do not penalize the silent target either. A
metric that credited correct realization would score the correct image higher
in both cases; the conventional metrics reward the disruptive
over-realization and are indifferent to the silent one.

\subsection{Prompt expansion: full results}
\label{app:expansion}

Table~\ref{tab:expansion} gives the outcome distribution of the five open
generators on the raw prompts and on the prompts rewritten by the Qwen-Image
rewriter and by Ideogram Magic Prompt, per family, with the change from the
raw prompt in parentheses. \S\ref{sec:expansion} discusses the
over-realization rates; Appendix~\ref{app:rewritetreatment} classifies what
each rewrite does with the target.

\begin{table}[t]
\centering
\small
\setlength{\tabcolsep}{3pt}
\scriptsize
\begin{adjustbox}{max width=\textwidth}
\resizebox{\textwidth}{!}{%
\begin{tabular}{l cccc@{\hspace{6pt}}cccc@{\hspace{6pt}}cccc@{\hspace{6pt}}cccc}
\toprule
& \multicolumn{4}{c}{\textbf{Existence-canceling}} & \multicolumn{4}{c}{\textbf{Mental-state}} & \multicolumn{4}{c}{\textbf{Figurative}} & \multicolumn{4}{c}{\textbf{Perspectival}} \\
\cmidrule(lr){2-5} \cmidrule(lr){6-9} \cmidrule(lr){10-13} \cmidrule(lr){14-17}
Generator + rewriter & DO{\tiny$\downarrow$} & SO{\tiny$\downarrow$} & I{\tiny$\uparrow$} & W{\tiny$\uparrow$} & DO{\tiny$\downarrow$} & SO{\tiny$\downarrow$} & I{\tiny$\uparrow$} & W{\tiny$\uparrow$} & DO{\tiny$\downarrow$} & SO{\tiny$\downarrow$} & I{\tiny$\uparrow$} & W{\tiny$\uparrow$} & DO{\tiny$\downarrow$} & SO{\tiny$\downarrow$} & I{\tiny$\uparrow$} & W{\tiny$\uparrow$} \\
\midrule
\emph{FLUX.1-dev} (raw) & .43 & .00 & .00 & .57 & .44 & .28 & .03 & .24 & .09 & .30 & \textbf{.14} & .46 & \textbf{.56} & .19 & .01 & \textbf{.24} \\
\quad + Qwen rewriter & .33 \dgood{(\textminus.10)} & .00 \dnc{(0)} & .00 \dnc{(0)} & .67 \dgood{(+.10)} & .41 \dgood{(\textminus.03)} & \textbf{.25} \dgood{(\textminus.03)} & \textbf{.05} \dgood{(+.02)} & .29 \dgood{(+.05)} & .05 \dgood{(\textminus.04)} & .21 \dgood{(\textminus.09)} & .09 \dbad{(\textminus.05)} & .65 \dgood{(+.19)} & .65 \dbad{(+.09)} & .20 \dbad{(+.01)} & .01 \dnc{(0)} & .14 \dbad{(\textminus.10)} \\
\quad + Ideogram Magic & \textbf{.11} \dgood{(\textminus.32)} & .00 \dnc{(0)} & .00 \dnc{(0)} & \textbf{.89} \dgood{(+.32)} & \textbf{.15} \dgood{(\textminus.29)} & .32 \dbad{(+.04)} & .04 \dgood{(+.01)} & \textbf{.49} \dgood{(+.25)} & \textbf{.00} \dgood{(\textminus.09)} & \textbf{.20} \dgood{(\textminus.10)} & .04 \dbad{(\textminus.10)} & \textbf{.76} \dgood{(+.30)} & .62 \dbad{(+.06)} & \textbf{.15} \dgood{(\textminus.04)} & .01 \dnc{(0)} & .22 \dbad{(\textminus.02)} \\
\addlinespace
\emph{Qwen-Image} (raw) & .64 & .00 & .00 & .36 & .66 & \textbf{.22} & .01 & .10 & .28 & .33 & .07 & .31 & .67 & .22 & \textbf{.01} & .10 \\
\quad + Qwen rewriter & .81 \dbad{(+.17)} & .00 \dnc{(0)} & .00 \dnc{(0)} & .19 \dbad{(\textminus.17)} & .50 \dgood{(\textminus.16)} & .35 \dbad{(+.13)} & \textbf{.10} \dgood{(+.09)} & .05 \dbad{(\textminus.05)} & .14 \dgood{(\textminus.14)} & .28 \dgood{(\textminus.05)} & \textbf{.12} \dgood{(+.05)} & .46 \dgood{(+.15)} & .72 \dbad{(+.05)} & .20 \dgood{(\textminus.02)} & .00 \dbad{(\textminus.01)} & .08 \dbad{(\textminus.02)} \\
\quad + Ideogram Magic & \textbf{.08} \dgood{(\textminus.56)} & .00 \dnc{(0)} & .00 \dnc{(0)} & \textbf{.92} \dgood{(+.56)} & \textbf{.26} \dgood{(\textminus.40)} & \textbf{.22} \dnc{(0)} & .06 \dgood{(+.05)} & \textbf{.46} \dgood{(+.36)} & \textbf{.04} \dgood{(\textminus.24)} & \textbf{.22} \dgood{(\textminus.11)} & .07 \dnc{(0)} & \textbf{.67} \dgood{(+.36)} & \textbf{.64} \dgood{(\textminus.03)} & \textbf{.19} \dgood{(\textminus.03)} & \textbf{.01} \dnc{(0)} & \textbf{.16} \dgood{(+.06)} \\
\addlinespace
\emph{OmniGen2} (raw) & .51 & .00 & .00 & .49 & .64 & \textbf{.22} & .03 & .11 & .18 & .26 & \textbf{.08} & .45 & \textbf{.63} & .16 & \textbf{.01} & .18 \\
\quad + Qwen rewriter & .20 \dgood{(\textminus.31)} & .00 \dnc{(0)} & .00 \dnc{(0)} & .80 \dgood{(+.31)} & .40 \dgood{(\textminus.24)} & .31 \dbad{(+.09)} & \textbf{.08} \dgood{(+.05)} & .21 \dgood{(+.10)} & .09 \dgood{(\textminus.09)} & .18 \dgood{(\textminus.08)} & .06 \dbad{(\textminus.02)} & .67 \dgood{(+.22)} & .66 \dbad{(+.03)} & .18 \dbad{(+.02)} & .00 \dbad{(\textminus.01)} & .16 \dbad{(\textminus.02)} \\
\quad + Ideogram Magic & \textbf{.09} \dgood{(\textminus.42)} & .00 \dnc{(0)} & .00 \dnc{(0)} & \textbf{.91} \dgood{(+.42)} & \textbf{.16} \dgood{(\textminus.48)} & .27 \dbad{(+.05)} & .07 \dgood{(+.04)} & \textbf{.50} \dgood{(+.39)} & \textbf{.02} \dgood{(\textminus.16)} & \textbf{.14} \dgood{(\textminus.12)} & .07 \dbad{(\textminus.01)} & \textbf{.76} \dgood{(+.31)} & \textbf{.63} \dnc{(0)} & \textbf{.14} \dgood{(\textminus.02)} & .00 \dbad{(\textminus.01)} & \textbf{.23} \dgood{(+.05)} \\
\addlinespace
\emph{SD3.5-Medium} (raw) & .34 & .00 & .00 & .66 & .66 & \textbf{.21} & .01 & .12 & .09 & .31 & .06 & .53 & .66 & \textbf{.11} & .00 & .22 \\
\quad + Qwen rewriter & .27 \dgood{(\textminus.07)} & .00 \dnc{(0)} & .00 \dnc{(0)} & .73 \dgood{(+.07)} & .47 \dgood{(\textminus.19)} & .22 \dbad{(+.01)} & \textbf{.05} \dgood{(+.04)} & .26 \dgood{(+.14)} & .09 \dnc{(0)} & .23 \dgood{(\textminus.08)} & \textbf{.12} \dgood{(+.06)} & .55 \dgood{(+.02)} & .68 \dbad{(+.02)} & .18 \dbad{(+.07)} & \textbf{.01} \dgood{(+.01)} & .11 \dbad{(\textminus.11)} \\
\quad + Ideogram Magic & \textbf{.07} \dgood{(\textminus.27)} & .00 \dnc{(0)} & .00 \dnc{(0)} & \textbf{.93} \dgood{(+.27)} & \textbf{.23} \dgood{(\textminus.43)} & .28 \dbad{(+.07)} & .00 \dbad{(\textminus.01)} & \textbf{.48} \dgood{(+.36)} & \textbf{.02} \dgood{(\textminus.07)} & \textbf{.19} \dgood{(\textminus.12)} & .04 \dbad{(\textminus.02)} & \textbf{.75} \dgood{(+.22)} & \textbf{.62} \dgood{(\textminus.04)} & \textbf{.11} \dnc{(0)} & .00 \dnc{(0)} & \textbf{.27} \dgood{(+.05)} \\
\addlinespace
\emph{SD3.5-Large} (raw) & .64 & .00 & .00 & .36 & .61 & \textbf{.23} & .03 & .13 & .20 & .40 & \textbf{.11} & .28 & \textbf{.58} & .16 & \textbf{.02} & \textbf{.22} \\
\quad + Qwen rewriter & .34 \dgood{(\textminus.30)} & .00 \dnc{(0)} & .00 \dnc{(0)} & .66 \dgood{(+.30)} & .43 \dgood{(\textminus.18)} & .30 \dbad{(+.07)} & \textbf{.04} \dgood{(+.01)} & .23 \dgood{(+.10)} & .07 \dgood{(\textminus.13)} & .30 \dgood{(\textminus.10)} & .09 \dbad{(\textminus.02)} & .54 \dgood{(+.26)} & .67 \dbad{(+.09)} & .22 \dbad{(+.06)} & .00 \dbad{(\textminus.02)} & .09 \dbad{(\textminus.13)} \\
\quad + Ideogram Magic & \textbf{.11} \dgood{(\textminus.53)} & .00 \dnc{(0)} & .00 \dnc{(0)} & \textbf{.89} \dgood{(+.53)} & \textbf{.27} \dgood{(\textminus.34)} & .24 \dbad{(+.01)} & .02 \dbad{(\textminus.01)} & \textbf{.47} \dgood{(+.34)} & \textbf{.03} \dgood{(\textminus.17)} & \textbf{.17} \dgood{(\textminus.23)} & .03 \dbad{(\textminus.08)} & \textbf{.77} \dgood{(+.49)} & .69 \dbad{(+.11)} & \textbf{.15} \dgood{(\textminus.01)} & .01 \dbad{(\textminus.01)} & .15 \dbad{(\textminus.07)} \\
\bottomrule
\end{tabular}}
\end{adjustbox}
\caption{Performance of prompt expansion. }
\label{tab:expansion}
\end{table}

\subsection{Experiment on the in-the-wild set}
\label{app:wild}

The curated prompts of OverReal-Gen are written or synthesized for the benchmark, and
the hand-written families are admitted only after a pilot model over-realized them
(\S\ref{sec:overreal-gen}). We also evaluate the systems on an in-the-wild set of prompts that users wrote for
their own purposes, selected on the wording of the prompt alone.

\paragraph{Sources and sampling}
We pool the prompts of DiffusionDB \citep{wang2022diffusiondb} and Pick-a-Pic
\citep{kirstain2023pickapic}, collapse whitespace, drop duplicates, drop DiffusionDB
prompts whose text or image carries an NSFW score of 0.5 or more, keep prompts of 3 to
120 words that are at least 90\% ASCII, and draw 7{,}000 DiffusionDB and 3{,}000
Pick-a-Pic prompts at random.

\paragraph{Classification}
Each of the 10{,}000 prompts is classified into one of the four families of
\S\ref{sec:families} or none, based on the cue words of each family. Of the 10{,}000 prompts, 202 (2.0\%)
qualify: 62 existence-canceling, 3 mental-state, 129 figurative and 8 perspectival. To obtain enough prompts of the two rare families, we add another 135 mental-state and 60 perspectival prompts from the rest of the datasets, also based on cue words.

\paragraph{Experiments}
The 397 prompts are generated under the settings of \S\ref{sec:setup}, one image per
prompt and system, and the images are labeled by the automatic annotator of
\S\ref{sec:autoannot}. Table~\ref{tab:mainresults}(b) gives the results, and the conclusions of \S\ref{sec:main} carry over. First, every system over-realizes in every family: pooled over all
systems, the over-realization rate is 0.60 on existence-canceling prompts, 0.28 on
figurative, 0.51 on mental-state and 0.80 on perspectival. Second, the form of the
failure follows the family: perspectival targets break
the scene (mostly DO), figurative targets slip in silently (almost entirely SO), and
mental-state targets sit between; the proprietary systems handle mental-state prompts
largely by integrated depiction (I of 0.58 to 0.64), which the open models rarely
manage. Third, the curated prompts run 0.1 to 0.3 higher than the in-the-wild prompts
in the mental-state and figurative families. Existence-canceling goes the other way:
negations in the wild mostly cancel an attribute of the subject (\emph{no nose},
\emph{faceless}, \emph{without laces}) rather than an object in the scene, and every
system, the proprietary ones included, over-realizes at least 28\% of them, far above
its rate on the templated prompts of \S\ref{sec:overreal-gen}.

\paragraph{\rev{Model rankings on the two sets}}
\rev{Ranking the eight systems by their over-realization rate, the Spearman
correlation between the two sets is 0.90 on existence-canceling and 0.83
on mental-state prompts, so the rankings there are stable. On figurative
and perspectival prompts it is near zero ($-0.26$ and $0.05$), but there
the systems lie close together on either set: in the wild, all eight fall
between 22\% and 32\% on figurative and between 74\% and 88\% on
perspectival prompts, so the ranking itself carries little information.
The form of the failure also agrees:
Table~\ref{tab:wilddoshare} gives the disruptive share of over-realization,
pooled over systems, and the two sets agree within five points except in
the figurative family. That gap has two causes. (1) By a keyword rule
checked by hand, about 90\% of the wild figurative prompts compare
appearance or shape (\emph{a cloud shaped like a snake}) and 10\% compare
a quality or behavior (\emph{as stubborn as a mule}); on OverReal-Gen the
split is about 15\% against 85\%. In an appearance comparison the target's
look is part of the request, so rendering it rarely breaks the scene; in a
quality comparison the target has no place in the scene, so rendering it
does. (2) The annotators of OverReal-Gen were asked not to reuse one
template, so that the prompts cover diverse failures, whereas the wild
figurative prompts are dominated by ``X shaped like Y'' art prompts.}

\begin{table}[!htb]
\centering
\color{blue}
\small
\begin{tabular}{l cc}
\toprule
& OverReal-Gen & In the wild \\
\midrule
Existence-canceling & 1.00 & 1.00 \\
Mental-state        & .69  & .66 \\
Figurative          & .35  & .08 \\
Perspectival        & .79  & .84 \\
\bottomrule
\end{tabular}
\caption{Disruptive share of over-realization, DO$/$(DO$+$SO), pooled over
the eight systems, on the raw prompts of OverReal-Gen and on the in-the-wild
prompts.}
\label{tab:wilddoshare}
\end{table}

\paragraph{\rev{Nano Banana and the screening}}
\rev{Since the hand-written prompts were screened with Gemini
(\S\ref{sec:overreal-gen}), we check whether the screening sets Nano Banana
apart from the other systems. Ranked by over-realization rate from highest
to lowest, Nano Banana is 6th, 7th, 5th and 1st on OverReal-Gen
(existence-canceling, mental-state, figurative, perspectival) and 7th, 7th,
7th and 6th in the wild. The screening changes its rank only on
perspectival prompts; on the other three families its rank is the same on
the two sets.}

\paragraph{\rev{Human check of the automatic labels}}
\rev{The in-the-wild rates rest on the automatic annotator alone, whose
agreement with the humans was measured on the curated images of
OverReal-Det (Table~\ref{tab:autoannot}). To check its labels on the wild
images, we drew a stratified sample of 202 of them, stratified by family
and automatic label with the perspectival family oversampled (100 images,
and 34 from each other family), and one of the authors relabeled them by
hand, blind to the automatic label. Table~\ref{tab:wildcheck} gives
Cohen's $\kappa$ between the two, with the values on OverReal-Det for
reference, and the over-realization rate after reweighting: each automatic
label is replaced by the distribution of human labels observed for it in
the sample. The automatic labels hold up. Over the four labels, the
agreement is 0.70 in the wild, the same as the two human annotators'
agreement with each other on OverReal-Det, and 0.65 on the perspectival
images, close to the human-human value of 0.75. The annotator over-calls
over-realization slightly: the reweighted rate is below the automatic one
in every family, by 2 to 12 points, but the pattern across families is
unchanged, and perspectival prompts remain far above the others at 68\%.}

\begin{table}[!htb]
\centering
\color{blue}
\footnotesize
\setlength{\tabcolsep}{4pt}
\resizebox{\textwidth}{!}{%
\begin{tabular}{l r ccc c cc}
\toprule
& & \multicolumn{3}{c}{$\kappa$, four labels} & $\kappa$, OR vs.\ correct & \multicolumn{2}{c}{OR rate in the wild} \\
\cmidrule(lr){3-5} \cmidrule(lr){6-6} \cmidrule(lr){7-8}
Family & $n$ & wild & Det, AI-human & Det, humans & wild & automatic & reweighted \\
\midrule
Existence-canceling & 34  & .82 & 1.00 & .83 & .82 & .60 & .51 \\
Mental-state        & 34  & .76 & .42  & .47 & .76 & .51 & .42 \\
Figurative          & 34  & .53 & .60  & .67 & .53 & .28 & .26 \\
Perspectival        & 100 & .65 & .41  & .75 & .69 & .80 & .68 \\
\midrule
All                 & 202 & .70 & .58  & .70 & .72 & --- & --- \\
\bottomrule
\end{tabular}}
\caption{Human check of the automatic labels in the wild. Cohen's $\kappa$
between the automatic annotator and a human annotator on a stratified
sample of $n$ wild images, over the four labels and over the collapse to
over-realized (OR) against correct, with the AI-human and human-human values
on OverReal-Det (Table~\ref{tab:autoannot}) for reference; and the
over-realization rate in the wild from the automatic labels and after
reweighting them by the human labels.}
\label{tab:wildcheck}
\end{table}

\paragraph{Conclusions}
First, apart from existence-canceling, the systems behave on
the prompts that users wrote much as they do on the prompts we wrote: the pattern
across families and systems is the same, and the rates differ by at most 0.3. The
curated prompts are harder, as intended, but they do not produce a failure that
natural prompts lack, so the results of \S\ref{sec:experiments} are not an artifact
of the curation. Second, the four situations account for about 2\% of the prompts in
current text-to-image datasets, so existing benchmarks under-represent them and have
not evaluated them. Mental-state and perspectival prompts are the rarest, under 0.1\%
each. We keep them in OverReal nonetheless: the same situations arise whenever a
prompt describes what a character thinks or what a viewpoint hides, which is routine
in narrative prompts and in video generation. A model that cannot keep such a target
out of a single image will not keep it out of a video either.

\subsection{How the rewriters treat the target}
\label{app:rewritetreatment}

For the analysis of \S\ref{sec:expansion} we read every rewritten prompt
and recorded what became of the target. Each
rewrite falls into one of four categories:
\begin{itemize}
\item \textbf{dropped}: the target is no longer mentioned, under any name;
\item \textbf{explicitly excluded}: the rewrite states that the target is
  absent or must not appear (``no keyboard is present'', ``the letter p is
  obscured by shadow'');
\item \textbf{mentioned}: the target is named but nothing in the rewrite
  asks for it to be shown, whether it keeps the status the original prompt
  gave it (``remembering the trumpet'', ``as fierce as a tiger'', ``inside the
  closed box is a teapot'') or not;
\item \textbf{explicitly depicted}: the rewrite describes the target as
  something to render in view (``a tiger stands in the corner'', ``visible
  through a gap in the lid'', ``the hidden side reveals a red tower'').
\end{itemize}
A first pass assigned the categories by rule, matching the target's words
against the rewrite and looking for negation and visibility phrases near
them; every rewrite that the rules flagged, every rewrite marked as dropped
that still produced an over-realized image, and a random sample of the rest
were then read and corrected by hand. Eight existence-canceling items whose
target is a normal part of the scene, such as a keyboard beside a laptop,
are left out of this analysis. Table~\ref{tab:rewritetreatment} gives the
distribution per family and rewriter, and Table~\ref{tab:rewritetreatmentor}
the over-realization rate of the corresponding images by category.

\begin{table}[h]
\centering
\small
\begin{tabular}{ll cccc}
\toprule
Family & Rewriter & Dropped & Excluded & Mentioned & Depicted \\
\midrule
Existence-canceling & Qwen & 0 & 98 & 2 & 0 \\
 & Ideogram & 95 & 2 & 2 & 0 \\
Mental-state & Qwen & 0 & 1 & 64 & 35 \\
 & Ideogram & 41 & 0 & 40 & 19 \\
Figurative & Qwen & 15 & 0 & 77 & 8 \\
 & Ideogram & 55 & 0 & 36 & 9 \\
Perspectival & Qwen & 0 & 1 & 42 & 56 \\
 & Ideogram & 10 & 2 & 44 & 45 \\
\bottomrule
\end{tabular}
\caption{How the rewriters treat the target, as a percentage of the rewritten prompts of each family: the target is dropped, explicitly excluded (``no keyboard is present''), mentioned, or explicitly to be depicted (``visible through a gap'').}
\label{tab:rewritetreatment}
\end{table}

\begin{table}[h]
\centering
\small
\begin{tabular}{l cccc}
\toprule
Family & Dropped & Excluded & Mentioned & Depicted \\
\midrule
Existence-canceling & 0 & 35 & 5 & --- \\
Mental-state & 4 & --- & 71 & 82 \\
Figurative & 4 & --- & 33 & 75 \\
Perspectival & 2 & 100 & 86 & 87 \\
\midrule
Overall & 2 & 38 & 60 & 84 \\
\bottomrule
\end{tabular}
\caption{Over-realization rate (percent) of the images generated from rewritten prompts, by how the rewrite treats the target, pooled over both rewriters and the five open generators. Cells with fewer than 10 images are left blank.}
\label{tab:rewritetreatmentor}
\end{table}

\subsection{Text encoder}
\label{app:encoder}

The backbone never sees the prompt, only the encoder's output, so we ask
whether that output still distinguishes the original prompt from one that
calls for the target. Both experiments below use the plain-mention controls
of \S\ref{sec:attention}: in a control, the words that keep the target out
of the scene are replaced by words that put it in, and the target word is
kept, so that \emph{``The coach is as fierce as a tiger''} becomes
\emph{``The coach is standing next to a tiger''} (one example per family in
Table~\ref{tab:controls}).

\begin{table}[t]
\centering
\footnotesize
\setlength{\tabcolsep}{4pt}
\begin{tabular}{l ccccc}
\toprule
Encoder & Existence-canceling & Mental-state & Figurative & Perspectival & All \\
\midrule
CLIP-L/14      & .95 $\pm$ .04 & .81 $\pm$ .06 & .87 $\pm$ .03 & .53 $\pm$ .02 & .79 $\pm$ .02 \\
CLIP-bigG/14   & .96 $\pm$ .03 & .87 $\pm$ .05 & .87 $\pm$ .03 & .60 $\pm$ .03 & .82 $\pm$ .02 \\
T5-XXL         & .95 $\pm$ .04 & .89 $\pm$ .04 & .94 $\pm$ .02 & .62 $\pm$ .03 & .85 $\pm$ .02 \\
Qwen2.5-VL-7B  & 1.00 $\pm$ .00 & .92 $\pm$ .04 & .91 $\pm$ .02 & .60 $\pm$ .03 & .86 $\pm$ .01 \\
\bottomrule
\end{tabular}
\caption{Accuracy of a linear probe that tells an original prompt from
its plain-mention control, from the contextual vector of the target word at
the layer each pipeline feeds to the backbone, $\pm$ half the width of the
bootstrap 95\% interval. Chance is 50\%.}
\label{tab:probe}
\end{table}

\begin{figure}[t]
\centering
\includegraphics[width=\textwidth]{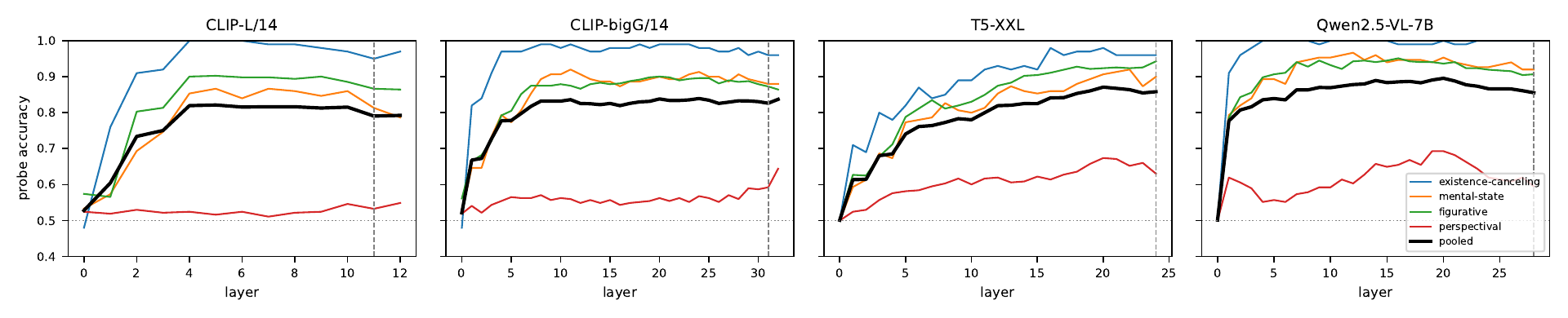}
\caption{Accuracy of the target-word probe of Table~\ref{tab:probe} at every
layer of each encoder, per family and pooled. Layer 0 is the embedding
output; the dashed line marks the layer the pipeline feeds to the backbone.
The perspectival family stays near chance at every layer of every encoder.}
\label{fig:sweep}
\end{figure}

\paragraph{Over-realization is not due to a lack of information in the encoder output, except on perspectival prompts.}
A linear probe is trained to tell the two prompts apart from the contextual
vector of the target word, at the layer the pipeline feeds to the backbone,
for the four encoders of the open models: CLIP-L and CLIP-bigG \citep{radford2021clip} (SD3.5),
T5-XXL \citep{raffel2020t5} (SD3.5 and FLUX) and Qwen2.5-VL-7B \citep{bai2025qwen25vl} (Qwen-Image). It is trained on
all four families together and tested with cross-validation grouped by
target word. On existence-canceling, mental-state and figurative prompts
the probe separates the two prompts at 81\% to 100\%, while on perspectival
prompts it stays close to chance, at 53\% to 62\%, at every layer
(Table~\ref{tab:probe}, Figure~\ref{fig:sweep}). Furthermore, within each of the three
families, the Spearman correlation between an item's probe margin and its
over-realization rate lies between $-0.28$ and $0.28$, with no consistent
sign across encoders. Thus, for existence-canceling, mental-state and figurative prompts, over-realization is not
explained by the encoding process; on perspectival
prompts the probe finds no linear evidence in the encoder output that the target is out of view.

\paragraph{Qwen-Image's encoder knows the target should not be visible, and the generator shows it anyway.}
Qwen-Image's encoder is also an instruction model, so we ask it directly:
\emph{``If this prompt were rendered as an image exactly as described, would
a $\langle$target$\rangle$ be visible in the image? Answer yes or no.''},
expecting \emph{no} for the original prompt and \emph{yes} for its control.
It answers correctly on 100\%, 75\% and 84\% of the existence-canceling,
mental-state and figurative prompts and on 56\% of the perspectival ones
(Table~\ref{tab:askencoder}); the same weights then condition images in
which the target appears in more than half of the cases
(Table~\ref{tab:mainresults}). This is another piece of evidence that the
information survives the encoding process for existence-canceling,
mental-state and figurative prompts; for the perspectival family we find no
evidence that it does.

\begin{table}[t]
\centering
\small
\begin{tabular}{l ccc}
\toprule
Family & Original prompt (no) & Plain-mention control (yes) & Both \\
\midrule
Existence-canceling & 1.00 & .76 & .88 \\
Mental-state        & .75  & 1.00 & .87 \\
Figurative          & .84  & .93 & .89 \\
Perspectival        & .56  & .85 & .70 \\
\bottomrule
\end{tabular}
\caption{Accuracy of Qwen2.5-VL-7B, the text encoder of Qwen-Image, when
asked directly whether the target would be visible.}
\label{tab:askencoder}
\end{table}

\subsection{Backbone attention: per family}
\label{app:attn}

\begin{table}[t]
\centering
\small
\resizebox{\textwidth}{!}{%
\begin{tabular}{l p{5.2cm} p{5.2cm}}
\toprule
Family & Original prompt & Plain-mention control \\
\midrule
Existence-canceling & a cat sitting on a couch with a dog on the floor, with no \textbf{book} in sight & a cat sitting on a couch with a dog on the floor, with a \textbf{book} in sight \\
Mental-state & Tim sits alone at a classroom desk, remembering the toy \textbf{robot} he lost years ago. & Tim sits alone at a classroom desk, holding the toy \textbf{robot} he got years ago. \\
Figurative & The coach is as fierce as a \textbf{tiger}. & The coach is standing next to a \textbf{tiger}. \\
Perspectival & A desk drawer is completely closed. Inside is a green \textbf{frog} toy. & A desk drawer is pulled open. Inside is a green \textbf{frog} toy. \\
\bottomrule
\end{tabular}}
\caption{Plain-mention controls: one example per family. The target word is
kept; the words that keep it out of the scene are replaced by words that put
it in.}
\label{tab:controls}
\end{table}

Table~\ref{tab:attnfam} gives the attention shares of \S\ref{sec:attention}
per family. Pairs per family: 100, 150, 472 and 368 for
existence-canceling, mental-state, figurative and perspectival prompts;
every value pools the two seeds.

\begin{table}[t]
\centering
\scriptsize
\setlength{\tabcolsep}{2.5pt}
\resizebox{\textwidth}{!}{%
\begin{tabular}{ll ccc@{\hspace{8pt}}ccc@{\hspace{8pt}}ccc}
\toprule
& & \multicolumn{3}{c}{\textbf{Original prompt}} & \multicolumn{3}{c}{\textbf{Plain-mention control}} & \multicolumn{3}{c}{\textbf{Ratios}} \\
\cmidrule(lr){3-5} \cmidrule(lr){6-8} \cmidrule(lr){9-11}
Model & Family & target $a_\mathrm{t}$ & cue $a_\mathrm{c}$ & other $a_\mathrm{o}$ & target $a'_\mathrm{t}$ & replacing $a'_\mathrm{r}$ & other $a'_\mathrm{o}$ & $\rho_\mathrm{t} = a_\mathrm{t}/a'_\mathrm{t}$ & $\rho_\mathrm{c} = a_\mathrm{c}/a'_\mathrm{r}$ & $\gamma = a_\mathrm{c}/a_\mathrm{o}$ \\
\midrule
FLUX.1-dev & Existence-canceling & 4.0 & 4.0 & 5.6 & 5.1 & 3.8 & 5.4 & .79 & 1.15 & .71 \\
           & Mental-state        & 3.8 & 3.0 & 4.2 & 4.8 & 3.5 & 4.2 & .74 & .87 & .73 \\
           & Figurative          & 5.6 & 4.7 & 6.1 & 6.5 & 5.6 & 5.7 & .83 & .80 & .77 \\
           & Perspectival        & 3.6 & 3.7 & 4.4 & 3.7 & 3.6 & 4.5 & .98 & 1.04 & .83 \\
           & \emph{All families} & 4.2 & 3.8 & 5.1 & 5.0 & 4.1 & 4.9 & .83 & .97 & .76 \\
\midrule
SD3.5-Large & Existence-canceling & 2.4 & 1.4 & 3.1 & 2.8 & 1.8 & 3.0 & .85 & .82 & .46 \\
            & Mental-state        & 2.4 & 1.3 & 2.3 & 2.6 & 1.6 & 2.3 & .85 & .79 & .54 \\
            & Figurative          & 4.2 & 2.1 & 3.3 & 4.4 & 2.6 & 3.1 & .92 & .80 & .63 \\
            & Perspectival        & 2.1 & 1.9 & 2.4 & 2.1 & 2.0 & 2.4 & .99 & .97 & .81 \\
            & \emph{All families} & 2.8 & 1.7 & 2.8 & 3.0 & 2.0 & 2.7 & .90 & .84 & .60 \\
\midrule
Qwen-Image & Existence-canceling & 5.2 & 4.6 & 6.3 & 6.3 & 6.2 & 6.3 & .80 & .84 & .73 \\
           & Mental-state        & 3.7 & 2.7 & 5.1 & 4.0 & 3.3 & 5.0 & .90 & .83 & .54 \\
           & Figurative          & 5.1 & 4.2 & 7.8 & 5.4 & 4.8 & 7.7 & .92 & .86 & .54 \\
           & Perspectival        & 3.6 & 3.4 & 5.2 & 3.6 & 3.5 & 5.2 & .97 & 1.03 & .66 \\
           & \emph{All families} & 4.4 & 3.8 & 6.1 & 4.8 & 4.4 & 6.0 & .90 & .89 & .62 \\
\bottomrule
\end{tabular}}
\caption{Attention in the backbone, per family and averaged over the four
families. $a$: percentage of the image-to-text attention per token of the
target words (t), the cue words (c) and the other words (o) under the
original prompt; $a'$: the same under the plain-mention control, where the
cue words are replaced (r). The ratios pair the original prompt and its
control under the same seed.}
\label{tab:attnfam}
\end{table}

\paragraph{The cue words lower the target's attention only slightly, and not at all on perspectival prompts.}
On existence-canceling, mental-state and figurative prompts the target
receives less attention under the original prompt than under the control in
every model ($\rho_\mathrm{t}$ 0.74 to 0.92). On perspectival prompts
$\rho_\mathrm{t}$ is 0.97 to 0.99: the cue words make almost no difference,
consistent with Appendix~\ref{app:encoder}, where the perspectival
information is already lost in the encoder.

\subsection{Backbone attention under prompt expansion}
\label{app:attnexpansion}

Expanded prompts are about five times longer than the original ones, so for
them we report the relative attention $r$: each token's share divided by
the prompt's average share.
Table~\ref{tab:attnexp} gives the target's $r$ under the original prompt
and under the two expansions of \S\ref{sec:expansion}, over the expanded
prompts in which the target survives verbatim (both seeds).

\begin{table}[t]
\centering
\scriptsize
\setlength{\tabcolsep}{4pt}
\begin{tabular}{l cc ccc@{\hspace{8pt}}ccc@{\hspace{8pt}}ccc}
\toprule
& & & \multicolumn{3}{c}{\textbf{FLUX.1-dev}} & \multicolumn{3}{c}{\textbf{SD3.5-Large}} & \multicolumn{3}{c}{\textbf{Qwen-Image}} \\
\cmidrule(lr){4-6} \cmidrule(lr){7-9} \cmidrule(lr){10-12}
Family & $n$ Qwen & $n$ Ideo. & orig. & Qwen & Ideo. & orig. & Qwen & Ideo. & orig. & Qwen & Ideo. \\
\midrule
Existence-canceling & 100 & 2   & .77  & 1.05 & ---  & .85  & .99  & ---  & .82 & .58  & ---  \\
Mental-state        & 136 & 76  & .96  & 1.60 & 1.85 & 1.13 & 1.05 & 1.42 & .81 & 1.32 & 1.05 \\
Figurative          & 398 & 196 & 1.00 & 1.37 & 1.35 & 1.40 & 1.09 & 1.14 & .78 & 1.43 & 1.17 \\
Perspectival        & 282 & 232 & .87  & 1.30 & 1.15 & .94  & 1.00 & 1.06 & .74 & 1.14 & 1.01 \\
\midrule
\emph{All families} & 916 & 506 & .90  & 1.33 & 1.45 & 1.08 & 1.04 & 1.21 & .79 & 1.12 & 1.08 \\
\bottomrule
\end{tabular}
\caption{Relative attention $r$ of the target words under the original
prompt (orig.) and under the expanded prompts of the Qwen rewriter and of
Ideogram Magic Prompt (Ideo.), per family and over all families; $n$ is the
number of expanded images per family and rewriter. Ideogram Magic Prompt
keeps the target in only two existence-canceling prompts, so that cell is
left out and the Ideogram mean excludes the family.}
\label{tab:attnexp}
\end{table}

\begin{figure}[t]
\centering
\includegraphics[width=\textwidth]{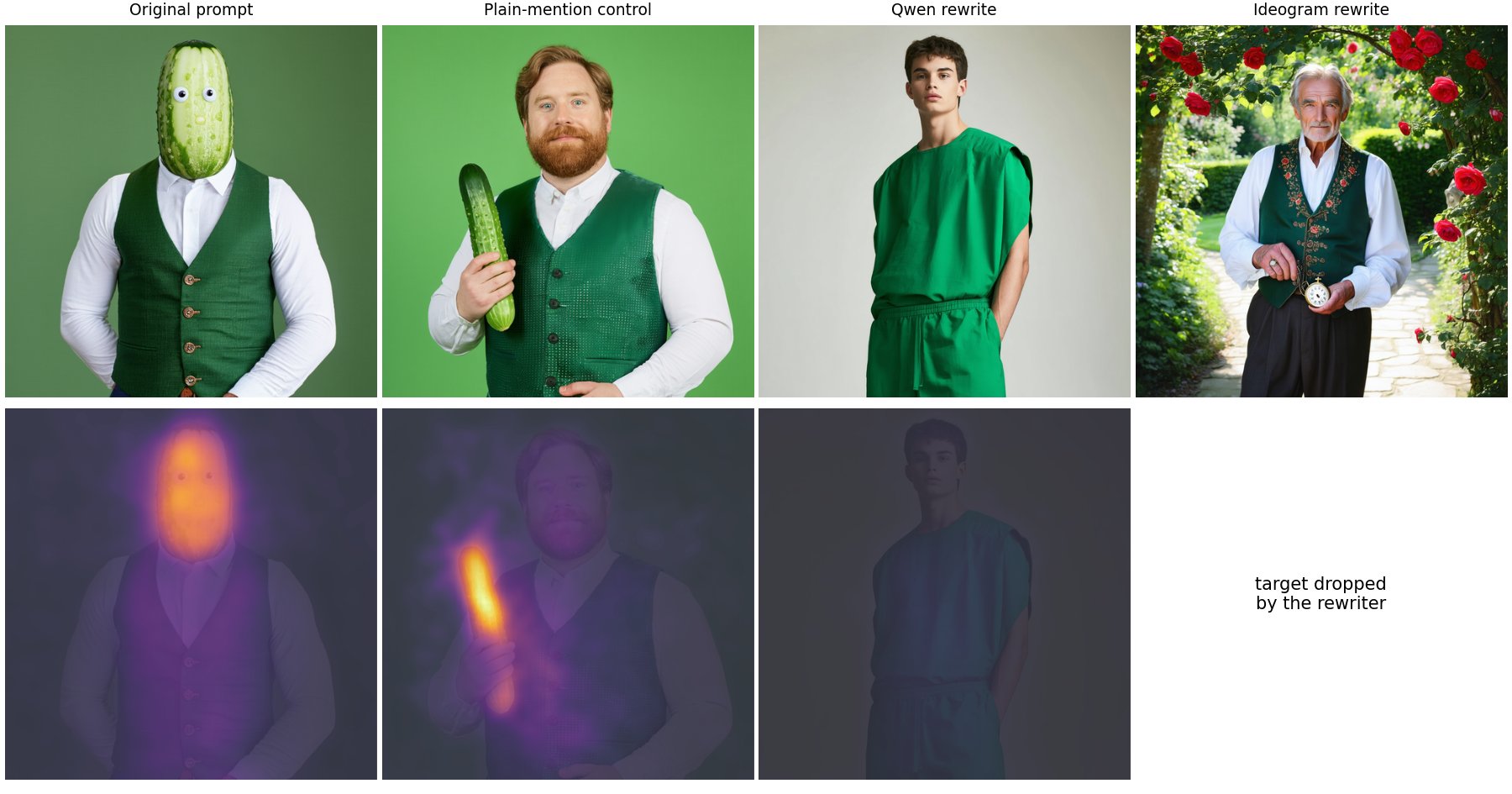}\\[4pt]
\includegraphics[width=\textwidth]{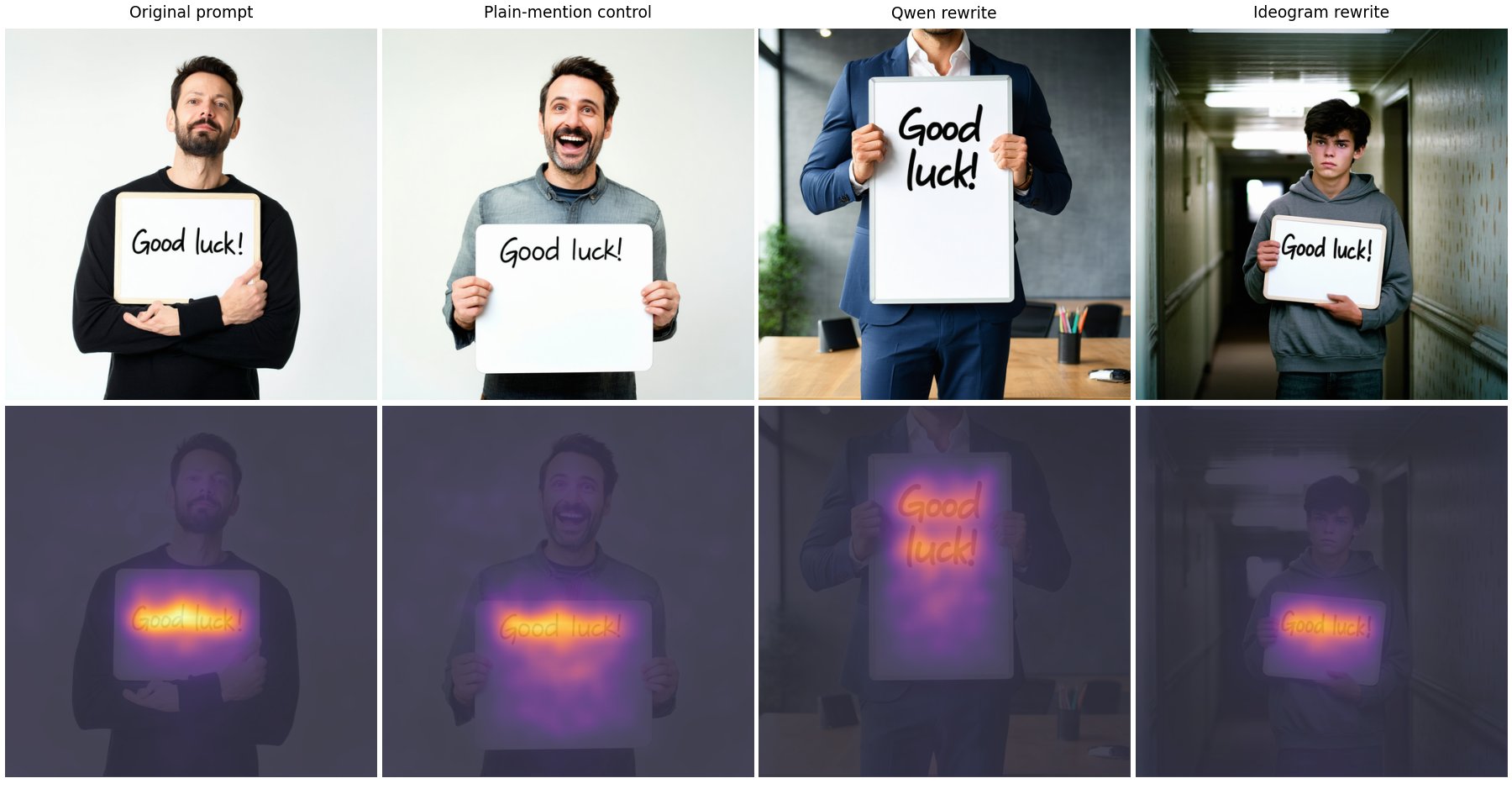}
\caption{Two items of Table~\ref{tab:taxonomy} under four prompts, SD3.5-Large:
the original prompt, its plain-mention control, the Qwen rewrite and the
Ideogram Magic Prompt rewrite (top row: the image; bottom row: attention from
the image tokens to the target words in the first denoising steps). Top: the
cucumber. The Qwen rewrite keeps \emph{cucumber} only as a color
(\emph{``the deep, fresh hue of a ripe cucumber''}); the backbone draws no
cucumber and the map stays flat. The Ideogram rewrite drops the word, so
there is no map. Bottom: the whiteboard whose \emph{``Good luck!''} faces
away from the camera. Both rewriters keep the words, both images show them,
and under all four prompts the attention sits on the writing.}
\label{fig:attnfour}
\end{figure}

\paragraph{Expansion does not lower the target's attention.}
Under the expanded prompts in which the rewriter kept the target, the
target receives at least as much attention as an average word in every
model and for both rewriters ($r$ = 1.04 to 1.45). The one exception is
Qwen-Image on existence-canceling prompts under the Qwen rewriter (0.58):
the rewriter restates the cancellation as a clause of its own, such as
\emph{No toilet is visible in the frame}, and inside that clause Qwen-Image
gives the target about half the attention of an ordinary word, while
FLUX.1-dev and SD3.5-Large read it like any other word (1.05 and 0.99).
Figure~\ref{fig:attnfour} shows two items under the original prompt, the
control and the two rewrites.

\subsection{\texorpdfstring{\rev{Explicit constraint in the prompt}}{Explicit constraint in the prompt}}
\label{app:constraint}

\rev{This experiment tests whether a generator withholds the target when the
prompt says so explicitly. We append the sentence \emph{``No \{target\} should appear in the image.''}
to all 75 mental-state prompts and to a random half (118) of the figurative
prompts of OverReal-Gen, and generate one image per prompt (seed 0) with
GPT-Image-1.5 and the five open-weight generators. The automatic annotator
labels these images against the original prompt, so the labels keep their
usual meaning. As the baseline we take the images of the same prompts in
the labeled sample of \S\ref{sec:autoannot} (raw prompts, both seeds).
GPT-Image declined eight of the constrained prompts (five mental-state,
three figurative), which are left out. Table~\ref{tab:constraint} gives
the outcome distribution per generator, and pooled over the five open-weight
generators.}

\begin{table}[!htb]
\centering
\color{blue}
\scriptsize
\setlength{\tabcolsep}{4pt}
\begin{tabular}{l rccccc@{\hspace{10pt}}rccccc}
\toprule
& \multicolumn{6}{c}{\textbf{Mental-state}} & \multicolumn{6}{c}{\textbf{Figurative}} \\
\cmidrule(lr){2-7} \cmidrule(lr){8-13}
Generator + prompt & $n$ & Target & DO & SO & I & W & $n$ & Target & DO & SO & I & W \\
\midrule
\emph{GPT-Image-1.5} (raw) & 100 & .89 & .35 & .26 & .28 & .11 & 42 & .57 & .14 & .26 & .17 & .43 \\
\quad + constraint & 70 & .14 & .01 & .03 & .10 & .86 & 115 & .08 & .00 & .05 & .03 & .92 \\
\addlinespace
\emph{FLUX.1-dev} (raw) & 99 & .76 & .44 & .28 & .03 & .24 & 47 & .47 & .09 & .23 & .15 & .53 \\
\quad + constraint & 74 & .77 & .50 & .23 & .04 & .23 & 116 & .59 & .15 & .37 & .07 & .41 \\
\addlinespace
\emph{Qwen-Image} (raw) & 99 & .90 & .67 & .22 & .01 & .10 & 45 & .69 & .27 & .38 & .04 & .31 \\
\quad + constraint & 75 & .91 & .72 & .19 & .00 & .09 & 118 & .76 & .26 & .43 & .07 & .24 \\
\addlinespace
\emph{OmniGen2} (raw) & 100 & .89 & .64 & .22 & .03 & .11 & 51 & .51 & .12 & .31 & .08 & .49 \\
\quad + constraint & 75 & .83 & .60 & .20 & .03 & .17 & 114 & .61 & .22 & .32 & .07 & .39 \\
\addlinespace
\emph{SD3.5-Medium} (raw) & 100 & .88 & .66 & .21 & .01 & .12 & 60 & .40 & .08 & .25 & .07 & .60 \\
\quad + constraint & 75 & .87 & .44 & .43 & .00 & .13 & 116 & .59 & .17 & .35 & .06 & .41 \\
\addlinespace
\emph{SD3.5-Large} (raw) & 100 & .87 & .61 & .23 & .03 & .13 & 53 & .68 & .17 & .38 & .13 & .32 \\
\quad + constraint & 75 & .87 & .68 & .19 & .00 & .13 & 117 & .74 & .25 & .39 & .10 & .26 \\
\midrule
\emph{Open-weight, pooled} (raw) & 498 & .86 & .60 & .23 & .02 & .14 & 256 & .54 & .14 & .31 & .09 & .46 \\
\quad + constraint & 374 & .85 & .59 & .25 & .01 & .15 & 581 & .66 & .21 & .37 & .07 & .34 \\
\bottomrule
\end{tabular}
\caption{Outcome distribution on the raw prompts and on the same prompts
with the explicit constraint \emph{``No \{target\} should appear in the
image.''} appended, for the mental-state prompts and a random half of the
figurative prompts of OverReal-Gen. Target is the share of images in which
the target appears (DO + SO + I); DO disruptive, SO silent, I integrated,
W withheld. The raw rows are over the labeled sample of
\S\ref{sec:autoannot} (both seeds), the constraint rows over one image per
prompt (seed 0).}
\label{tab:constraint}
\end{table}

\paragraph{\rev{GPT-Image follows the constraint; the open-weight generators do not.}}
\rev{With the constraint, the target almost disappears from GPT-Image's output:
it appears in 14\% of the mental-state images instead of 89\%, and in 8\%
of the figurative images instead of 57\%. For the five open-weight
generators the target appears as often as before in the mental-state family
(85\% against 86\%) and more often in the figurative family (66\% against
54\%). This matches their rates on the existence-canceling prompts, which
are explicit negations by construction: the open-weight generators
over-realize 34\% to 64\% of them (Table~\ref{tab:mainresults}). So
GPT-Image is able to exclude the target when told to. On the original
prompt it tends to treat the mentioned target almost as a requested one
and renders it, unless the prompt rules it out explicitly.}

\paragraph{\rev{An explicit constraint is not a general fix.}}
\rev{A prompt contains several mentions, any of which the model may
over-realize (Appendix~\ref{app:unexpected}), and in practice a user does
not say which mention to withhold. The fix should come from a model that
infers which mentions belong in the scene, not from a prompt that spells
it out.}

\section{Full Prompts and Annotation Guidelines}
\label{app:prompts}

\subsection{Automatic annotation protocol}
\label{app:autoannot}
Each image is scored by a cascade of yes/no questions, one API call per
question, with the image, the question and, from the second question on, the
original prompt. \texttt{T} stands for the target; the article is adjusted to
the target. Every family starts with the same screening question, ``Is this
image severely corrupted, or unrelated to the scene the prompt describes?'',
whose yes-answer gives \emph{other}.

\begin{description}
\item[Existence-canceling] Q1: Is there a T anywhere in this image?
  (yes: disruptive; no: withheld)
\item[Mental-state] Q1: Is there a T anywhere in this image? (no: withheld)
  Q2: In this image, is a T physically present in the scene, as an actual
  thing that is there with the person? (no: Q3)
  Q2b: In this image, is the T right beside the person and prominent in the
  picture, as opposed to distant, small or in the background?
  (yes: disruptive; no: silent)
  Q3: In this image, is a T shown as something the person imagines, remembers
  or believes, for example inside a thought bubble, a dream haze, a separate
  inset or a ghostly overlay? (yes: integrated; no: silent)
\item[Figurative] Q1: Is there a T anywhere in this image? (no: withheld)
  Q2: In this image, does the presence of a T make the scene absurd,
  impossible or nonsensical as a picture of what the prompt describes?
  (yes: disruptive)
  Q3: In this image, is a T merged with the person or thing it describes, as a
  deliberate visual metaphor in a stylized or poster-like composition?
  (yes: integrated; no: silent)
\item[Perspectival, hidden object] Q1: Is there a T anywhere in this image, in
  any form, including as a print, picture, sticker, logo or drawing?
  (no: withheld)
  Q2: According to the prompt, the T should be hidden from the viewer: inside,
  behind, underneath, or on the far side of something. In this image, is the T
  nevertheless exposed to the viewer at that very spot, for example because
  the container, door, curtain, barrier or surface is shown open, transparent,
  removed, or turned the wrong way? (yes: disruptive)
  Q3: In this image, is the T shown only indirectly, through a reflection, an
  inset, a thought bubble, or a picture taken from the in-scene person's own
  point of view? (yes: integrated; no: silent)
\item[Perspectival, hidden text] Q1: In this image, is the written side of the
  item that carries ``T'' (the letter, note, page, screen, card or sign the
  prompt describes) turned toward the viewer, so that its writing can be seen,
  whether or not it is legible? (no: withheld)
  Q2: In this image, is the writing ``T'' shown to the viewer only through a
  reflection, an inset, a thought bubble, or a picture taken from the in-scene
  person's own point of view? (yes: integrated)
  Q3: According to the prompt, the writing ``T'' should be out of the viewer's
  sight. In this image, is it nevertheless in plain view, on the very item the
  prompt describes? (yes: disruptive; no: silent)
\end{description}

\subsection{Annotation guide}
\label{app:guide}
Annotators labeled through a web interface that showed the image, the prompt,
and the target, together with the guide below (reproduced verbatim); each
person could pick one or two of the five labels per image.

\begin{itemize}
\item \textbf{disruptive}: the target appears, and its appearance causes at
least one of: (1) the image breaks common sense or logic, e.g.\ a reader
facing the back of the letter they are reading, a screen facing the camera
instead of its user, a tiger on a sports field; (2) the image contradicts the
prompt, e.g.\ the prompt says the target is absent, or sealed inside an opaque
box, yet it is visible.
\item \textbf{silent}: the target appears; unexpected, but not disruptive,
typically as background or a small detail. E.g.\ ``her face is like a red
apple'' shown as a red-faced child eating an apple; ``the roof is like a
banana'' with banana trees in the background.
\item \textbf{integrated}: the target appears, but in a natural way, within
what a normal viewer would accept. E.g.\ ``the roof is like a banana'' as a
banana-shaped roof and no banana trees; a thought bubble depicting what a
character imagines; contents shown inside a closed drawer as such; a poster
style that merges a person with their attribute.
\item \textbf{withheld}: the target does not appear, and the image does not
contradict the prompt. E.g.\ something in a closed bag shown only as the
closed bag; ``her face is like a red apple'' as just a red-faced child.
\item \textbf{other}: the image is severely corrupted, or does not match the
prompt at all.
\end{itemize}

\noindent You may select two labels, but please prefer a single label; pick
two only when the case is genuinely ambiguous.

\end{document}